\documentclass[runningheads]{llncs}
\usepackage{graphicx}
\usepackage{amsmath}
\usepackage{amssymb}
\usepackage{booktabs}
\usepackage{multirow}
\usepackage{algorithm} 
\usepackage{algorithmic} 
\usepackage{caption}
\usepackage{longtable, booktabs,etoolbox}
\usepackage{tikz}
\usepackage{comment}
\usepackage{amsmath,amssymb} 
\usepackage{color}
\usepackage[export]{adjustbox}
\usepackage{cite}
\usepackage[capitalize]{cleveref}
\crefname{section}{Sec.}{Secs.}
\Crefname{section}{Section}{Sections}
\Crefname{table}{Table}{Tables}
\crefname{table}{Tab.}{Tabs.}

\usepackage[T1]{fontenc}
\usepackage{graphicx}
\usepackage{lmodern}
\rmfamily
\DeclareFontShape{T1}{lmr}{bx}{sc}{<-> cmr10}{}
\DeclareFontFamily{OML}{zlmm}{}
\DeclareFontShape{OML}{zlmm}{m}{it}{<-> lmmi10}{}
\DeclareFontShape{OML}{zlmm}{b}{it}{<->ssub * zlmm/m/it}{}
\DeclareFontShape{OML}{zlmm}{bx}{it}{<->ssub * zlmm/m/it}{}
\DeclareMathVersion{Tinyb}
\SetSymbolFont{operators}{Tinyb}{T1}{lmr}{bx}{sc}
\SetSymbolFont{letters}{Tinyb}{OML}{zlmm}{m}{it}
\newenvironment{tinyb}{\bgroup\tiny\bfseries\scshape\mathversion{Tinyb}}{\egroup} 

\usepackage[accsupp]{axessibility}  
\makeatletter

\newcommand{\printfnsymbol}[1]{%
  \textsuperscript{\@fnsymbol{#1}}%
}
\makeatother

\usepackage[width=122mm,left=12mm,paperwidth=146mm,height=193mm,top=12mm,paperheight=217mm]{geometry}

\begin{document}
\pagestyle{headings}
\mainmatter
\def\ECCVSubNumber{5365}  

\title{SegDiscover: Visual Concept Discovery via Unsupervised Semantic Segmentation} 

\titlerunning{SegDiscover}
%

\author{Haiyang Huang\thanks{equal contribution}
 \and
Zhi Chen\printfnsymbol{1} \and
Cynthia Rudin}

\authorrunning{H. Huang et al.}
%
\institute{Duke University, Durham, NC 27708, USA\\
\email{\{hyhuang, zhichen, cynthia\}@cs.duke.edu}}
\maketitle

\begin{abstract}
Visual concept discovery has long been deemed important to improve interpretability of neural networks, because a bank of semantically meaningful concepts would provide us with a starting point for building machine learning models that exhibit intelligible reasoning process. Previous methods have disadvantages: either they rely on labelled support sets that incorporate human biases for objects that are ``useful,'' or they fail to identify multiple concepts that occur within a single image. We reframe the concept discovery task as an unsupervised semantic segmentation problem, and present SegDiscover, a novel framework that discovers semantically meaningful visual concepts from imagery datasets with complex scenes \textit{without supervision}. Our method contains three important pieces: generating concept primitives from raw images, discovering concepts by clustering in the latent space of a self-supervised pretrained encoder, and concept refinement via neural network smoothing.
Experimental results provide evidence that our method can discover multiple concepts within a single image and outperforms state-of-the-art unsupervised methods on complex datasets such as Cityscapes and COCO-Stuff. 
Our method can be further used as a neural network explanation tool by comparing results obtained by different encoders. 
\end{abstract}

\section{Introduction}

When we see a natural image, we can determine that it is comprised of entities, i.e., \textit{concepts}, some of which we can name (e.g., sky, sand, water, furniture, person) and some of which had not thought of naming (e.g., the piece of wall just between a vent and a door, or a patterned tin ceiling tile). A long-standing important question of computer vision is whether a ``natural'' set of concepts can be extracted directly from natural images. In other words, rather than relying on humans to define concepts manually, which is labor intensive and biased towards concepts that humans find directly useful and easy to name (e.g., objects such as chairs or tables), the question is whether such concepts can be extracted from images \textit{without} human supervision. If such a \textit{concept base} could be found, it would be valuable to myriad computer vision tasks, potentially acting like a basis in which to represent the natural visual world. Finding such a concept base is also a fundamental problem of interpretable machine learning \cite{rudin2021interpretable}, because models that combine known concepts can be made easier to understand than models that combine pixels in inexplicable ways.

However, despite many years of effort dedicated to finding such a base of concepts 
\cite{kim2018interpretability,chen2020concept,koh2020concept,zhou2018interpretable,zhou2018interpreting,locatello2020object,schwettmann2021toward,mao2019neuro,shen2021closed},  
this goal has remained elusive. Some previous works \cite{mao2019neuro,locatello2020object} successfully recovered a set of concepts in simplistic ``block worlds,'' but the real world is massively more difficult. Much work has been dedicated to studying concepts in \textit{supervised} settings, either with labels of the concepts or labels of the main classification tasks, but these are limited by the selection of concepts that humans have labeled \cite{kim2018interpretability,chen2020concept,koh2020concept,zhou2018interpretable,zhou2018interpreting,schwettmann2021toward} or by the human biases in labels of the main classification tasks \cite{ghorbani2019towards}. 

 
\begin{figure}[t]
\begin{center}
\includegraphics[width=1\linewidth]{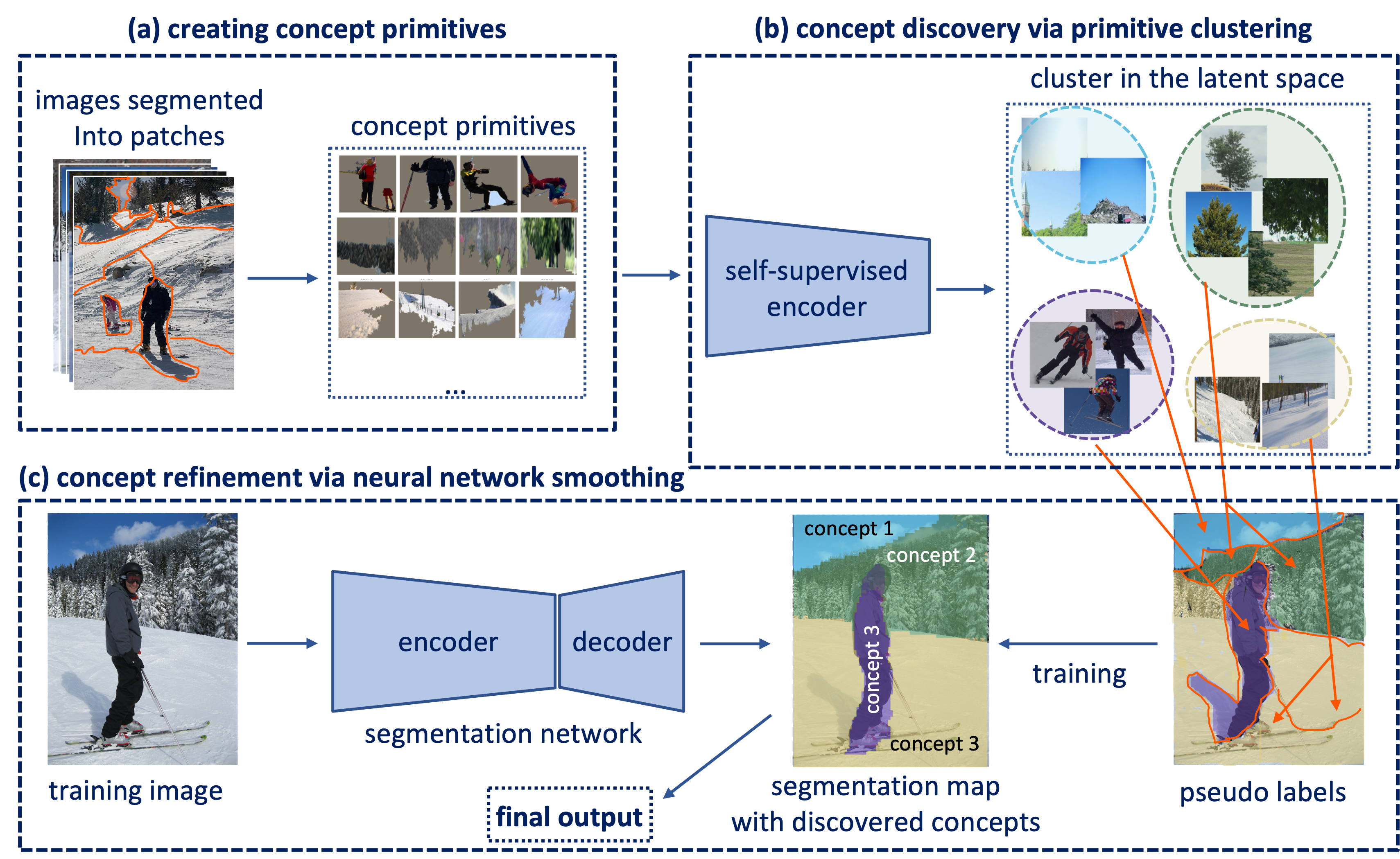}
   \caption{The SegDiscover concept discovery framework. The framework consists of three steps (a) \textit{creating concept primitives}, where images in the dataset are segmented into ``concept primitives'' (patches/super pixels); (b) \textit{concept discovery via primitive clustering}, where the primitive patches are feed into a pretrained self-supervised network, and potential concepts are discovered by clustering in its latent space; (c) \textit{concept refinement via neural network smoothing}, where the cluster labels are mapped back to the original images to form pseudo segmentation labels, and a segmentation network is trained to predict the pseudo labels given the original images. The segmentation network refine the learned concepts. The discovered concepts and segmentation maps are the outputs of SegDiscover.
   }
\label{fig:theme}
\vspace{-2.5em}
\end{center}
\end{figure}

Certainly, we cannot hope to discover concepts without some understanding of the natural world. In this work, we consider whether a few simple but powerful assumptions about how our visual world is comprised of concepts could bring us closer to the ultimate goal of creating a general concept base for it. Our assumptions are: 1) A natural image can be decomposed into a set of concepts, each element of which is represented by a connected patch of pixels. There can be multiple patches of the same concept in one image. For instance, an image of a room can be composed of doors, furniture, walls, etc. Visual concepts are determined by both 2) local visual similarity  (e.g., patches of sky look similar between images) and 3) global contextual information (e.g., patches of sky look similar to swimming pool but they are different contextually).


While these assumptions seem reasonable to us, we note that previous works do not always embody these assumptions and are often at odds with them: \cite {cho2021picie} does not explicitly utilize the first assumption, and thus learns pixel-level representations rather than patch-level representations; 
\cite{hwang2019segsort,zhang2020self,van2021unsupervised} focuses only on salient objects in the foreground but ignore important concepts in the background, which contradicts the first assumption; \cite{hwang2019segsort,ghorbani2019towards} do not incorporate any global contextual information to learn visual concepts. 
 
In this work, we directly embody the assumptions above within an algorithmic framework in order to discover concepts from natural images without any supervision. We do this as follows: To handle Assumption 1, we incorporate ideas from unsupervised image segmentation into our approach. To handle Assumption 2, we incorporate self-supervised neural networks and leverage the fact that patches of the same concept tend to cluster in the latent space. To handle Assumption 3, we refine the discovered concepts by designing a neural network training process to incorporate global information.
By basing our methodology only on these three assumptions, and not on labels provided by human experts, we ensure that our discovered concepts are not biased towards only those that humans might tend to label in supervised databases.

The resulting algorithm, called \textit{SegDiscover}, is a significant step forward towards unsupervised concept discovery in complex visual settings. Importantly, it is able to find coherent and meaningful visual concepts. Figure~\ref{fig:theme} shows how SegDiscover is able to fully segment an image and assign each piece of it to a concept class, all without any supervision. To evaluate SegDiscover, we considered its unsupervised semantic segmentation results on two complex real world datasets, where it outperforms state-of-the-art methods under various settings. Besides semantic segmentation, a further use-case for SegDiscover is as a neural network explanation tool: by replacing SegDiscover's encoder with the user's desired pre-trained encoder, SegDiscover's discovered concepts act as an explanation tool to assist in visualizing the latent space of a pretrained network.

\section{Related Work}
\label{sec:related}
Our work is closely related to concept-based explainability and interpretability, unsupervised semantic segmentation, and self-supervised learning. Table \ref{tab:comparison} summarizes the differences between the proposed method and related work. Our method distinguishes itself by considering unsupervised concept discovery and semantic segmentation \textit{simultaneously to inform each other}; it also explains behavior of the network in a way that combines pixel attributions on the input image with identifying what concepts are learned in the latent space. 
\begin{table}[t]
\caption{Comparison between the proposed method and related work.}
\begin{tinyb}
\begin{scriptsize}
\centering

\begin{tabular}{l|ccccc}
\hline
Method & \begin{tabular}[c]{@{}c@{}}Unsupervised\\ (no class label)\end{tabular} & \begin{tabular}[c]{@{}c@{}}Concept \\ Discovery\end{tabular} & \begin{tabular}[c]{@{}c@{}}Semantic\\ Segmentation\end{tabular} & \begin{tabular}[c]{@{}c@{}}Pixel\\ Attribution\end{tabular} & \begin{tabular}[c]{@{}c@{}}Explain \\ Latent Space\end{tabular} \\ \hline
Saliency-based Explanation &  &  &  & $\checkmark$ &  \\
Concept-based Explanation (supervised) &  &  &  &  & $\checkmark$ \\
Concept-based Explanation (unsupervised) &  & $\checkmark$ &  &  & $\checkmark$ \\
Unsupervised Semantic Segmentation & $\checkmark$ &  & $\checkmark$ &  &  \\
Self-supervised Networks & $\checkmark$ &  &  &  &  \\
SegDiscover (Our Method) & $\checkmark$ & $\checkmark$ & $\checkmark$ & $\checkmark$ & $\checkmark$ \\ \hline
\end{tabular}
\end{scriptsize}
\end{tinyb}
\label{tab:comparison}
\end{table}

\textbf{Concept-based Explainability and Interpretability.}
Interpretability of a neural network is often closely related to how its latent features operate. Believing that abstract concepts are learned in the hidden layers, post-hoc analyses have been proposed to measure the alignment of individual neurons with predefined concepts \cite{zhou2014object,zhou2018interpreting}.
While this is interesting, a majority of neurons have not been found to have an interpretable meaning, and it is unclear whether the discovered neurons truly represent all information about a concept. Instead of analyzing individual neurons, other post-hoc methods try to learn a linear combination of neurons to represent a predefined concept \cite{kim2018interpretability}, or even decompose the latent space into a basis defined by a large concept bank \cite{zhou2018interpretable}. All of these works use predefined concepts. Going beyond predefined concepts, recent work discovers new concepts by clustering patches in the latent space \cite{ghorbani2019towards}, and uses these unsupervised concepts to form a basis \cite{yeh2020completeness}. However, as shown by \cite{chen2020concept}, black-box networks trained on classification tasks can yield impure concepts. 

Alternatively, another line of work tries to build an inherently interpretable latent space made of predefined concepts \cite{chen2020concept,koh2020concept,losch2019interpretability,adel2018discovering}. 
However, these inherently interpretable methods require human annotation of concepts, which is what we want to eliminate in the proposed work.

\textbf{Unsupervised Semantic Segmentation.}
Semantic Segmentation had long been viewed as a problem that cannot be resolved in absence of human annotation -- until recently. A few attempts have been made to tackle the problem without supervision. IIC \cite{ji2019invariant} maximizes the mutual information of patch-level embeddings between different augmented views, but it largely relies on the hue of the pixels and does not work on more challenging cases. Some later works \cite{hwang2019segsort,zhang2020self,van2021unsupervised} utilize a pretrained contour or saliency detector as a segmentation prior to group pixels together and learn pixel-level embeddings that encode the semantic information. Recently, PiCIE \cite{cho2021picie} attempted unsupervised semantic segmentation on diverse scenes by clustering pixel-level embeddings between different augmented views. Our method instead leverages both pixel-level and patch-level information. PseudoSeg\cite{zou2020pseudoseg} utilized pseudo labels to improve segmentation results on low data regime. However, 
its pseudo label creation heavily relies on supervision signal from other datasets, whereas our method does not require \textit{any} labels. 


Although the task of unsupervised semantic segmentation could conceivably share a similar set of assumptions to those we use for concept discovery, the desiderata can be very different. In unsupervised semantic segmentation, the objective is to improve the quality of segmentation on \textit{predefined} classes and reduce the need for manual labeling. On the other hand, concept discovery aims to discover a set of visual concepts that \textit{may or may not appear} in the labels. Beyond performing segmentation, our work can also generate meaningful concept clusters (like Figure~\ref{fig:theme}b) and concept maps (like Figure~\ref{fig:theme}c second from right) 
that can be directly used for explanation for neural network behavior.



\textbf{Self-supervised Learning.}
In computer vision, self-supervised learning models are trained to carry out tasks such as predicting relative location of image parts \cite{1505.05192,1603.09246, IsolaZKA15, gidaris2018unsupervised}, colorization \cite{zhang2016colorful}, cluster assignments \cite{caron2018deep,caron2019unsupervised,1805.00385},
or distinguishing pictures from a set of irrelevant samples \cite{ExemplarNet,1711.06379,MoCo,Misra2019PILR,2002.05709,Mocov2,caron2020unsupervised,SimSiam}).
These networks have good embedding spaces that are useful in downstream tasks, such as image classification. 
We used self-supervised pretrained encoders to initialize the concept discovery process.


\section{Methods}

Given an image dataset $\mathcal{X}$, our goal is to discover a set of visual concepts $\mathcal{C}=\{1,2,\cdots,C\}$ that could best describe images in the dataset. Specifically, the visual concept discovery objective is formulated as an unsupervised semantic segmentation problem. For a given image $x\in \mathbb{R}^{H\times W \times 3}$ from dataset $\mathcal{X}$ (image size is $H\times W$ and it has 3 RGB channels), we want to learn a segmentation function $f_s \colon \mathbb{R}^{H\times W \times 3} \to \mathcal{C}^{H\times W}$, that can produce a segmentation map $f_s(x)$. The segmentation map is the same size as the image and identifies how image $x$ is composed of concepts in $\mathcal{C}$. Formulating concept discovery as a semantic segmentation problem emphasizes two points: (a) multiple concepts can co-occur in the same image, (b) the discovered concepts are complete, i.e., each pixel in $x$ is assigned to exactly one concept in $\mathcal{C}$.


However, since the entire task is \textit{unsupervised}, there is no obvious objective function $\mathcal{L}(\mathcal{X}, \mathcal{C}, f_s)$ to optimize. Our method relies on a self-supervised encoder $f_\Phi$. While self-supervised learning methods do not require labels, the networks learned by these methods are capable of learning meaningful representations that perform well for various downstream tasks \cite{Mocov2,2002.05709,caron2020unsupervised}. 
Thus, we envision $f_\Phi$ can place segments of the same concept near each other in the latent space, and segments of different concepts far from each other, so that the concepts can be easily extracted from the latent space.
In addition, self-supervised learning models usually have latent spaces that are transformation invariant, which is an important criterion that visual concepts should obey\cite{wu2018unsupervised}. 

Our framework is composed of three steps. First, raw images are decomposed into small patches containing potential visual concepts, called concept primitives (Figure~\ref{fig:theme}a, see Section \ref{sec:method:primitives}). Second, the concept primitives are fed into the self-supervised encoder $f_\Phi$ and we cluster their representations through an overclustering+reassignment clustering algorithm. The clusters form the discovered concepts set $\mathcal{C}$ (Figure~\ref{fig:theme}b, see Section \ref{sec:method:cluster}). Third, the clusters are used as pseudo segmentation labels to train a segmentation network (Figure~\ref{fig:theme}c, see Section \ref{sec:method:nn_smooth}). 
In the following sections, we will explain these steps in detail.


\subsection{Creating Concept Primitives}
\label{sec:method:primitives}



Semantic segmentation is usually decomposed into two separate subtasks: (1) dividing the image into segments with the same semantic meaning, and (2) predicting which class each segment belongs to. In the unsupervised setting, dividing the image into semantically meaningful segments is much harder than in the supervised setting. Previous works \cite{hwang2019segsort,zhang2020self,van2021unsupervised} either utilize saliency detection methods, which limit the number of segments (to two -- just foreground and background) and require additional supervision, or use contour detection methods that tend to severely oversegment the image. In real imagery datasets, especially in datasets with complex scenery, such methods are insufficient to define and separate the multitude of concepts that may appear in a single image.

To overcome these problems with unsupervised semantic segmentation, we first leverage traditional superpixel algorithms (e.g., \cite{felzenszwalb2004efficient}) to partition the whole image $\mathbf{x}_i$ into a set of segments at medium scale, i.e., local concept primitives, denoted as  $\{\mathbf{x}_i^{(j)}\}_{j=1}^{N_i}$, in which $N_i$ is the number of concept primitives in image $\mathbf{x}_i$. 
However, irregular shapes with unwanted boundaries will be generated (e.g., two pieces of the same bedroom wall, or parts of a single textured blanket being separated into parts), and these shapes can add noise to the concept discovery process. To solve this problem, we design a \textit{primitive-merging} approach to identify and merge these patches with their neighbors. The main steps are:

1. Finding the primitives that are either \textit{too small} (defined by area of the primitive divided by the area of the entire image) 
or \textit{too irregular} (defined by the perimeter divided by the square root of the area). 

2. Merging the primitive with its largest neighboring primitive that shares a similar hue.

A detailed description of the primitive-merging method is provided in the \textit{supplementary materials}. We provide an ablation study on how the primitive-merging affects the concept discovery quality in Section~\ref{sec:experiments}.

After this step, concept primitives are created, which provide a rough segmentation of the images. The concepts to which these primitives belong are unknown at this stage.


\subsection{Concept Discovery via Primitive Clustering} 
\label{sec:method:cluster}
Stepping back, we ask how to discover the concepts if we already have the perfect set of concept primitives. Ideally, we wish to have a function $f$ that maps all the primitives $\mathbf{x}_i^{(j)}$ into a space where primitives belonging to the same concepts are close to each other and those that are not are far apart. 
In such cases, a clustering algorithm would discover the concepts. 
The proposed method utilizes the fact that the latent representations in convolutional neural networks (CNNs) can effectively capture semantic meaning of images \cite{zhang2018unreasonable}. For $f$, we use a self-supervised encoder $f_\Phi$ (we try MoCo\cite{MoCo}, SwAV\cite{caron2020unsupervised}, DeepCluster\cite{caron2018deep}), pretrained on ImageNet images\cite{Imagenet} -- but not the labels. Note that self-supervised learning algorithms like \cite{MoCo,caron2018deep,caron2020unsupervised} \textit{do not use ImageNet labels}. 
We crop each primitive patch using the smallest box that bounds the patch, and resize it into an image. Areas not within the patch are set to the average color of the dataset.  We then feed the patches into the encoder $f_\Phi$ and get the latent representations of all the concept primitives and, as we next describe, we apply an overclustering+reassignment (OC-RA) clustering algorithm on these representations to generate $C$ clusters that represent the discovered set of visual concepts $\mathcal{C}$.

\begin{figure}[t]
  \centering
   \includegraphics[width=0.7\linewidth]{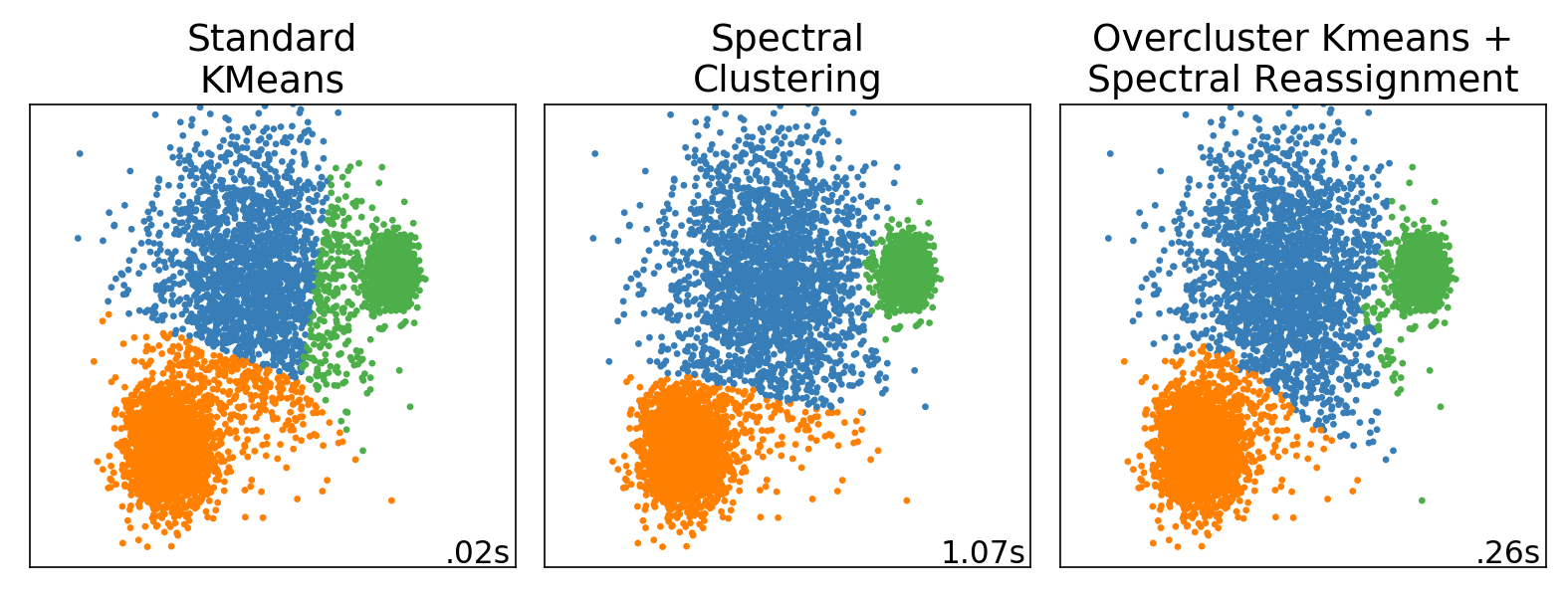}

  \caption{Effect of overclustering and reassignment (OC-RA) algorithm on simulated data.
  While the self-supervised encoder can place similar primitives together, identifying concepts from the space is non-trivial, since the radii of different concepts may vary significantly, thus k-means could fail. The OC-RA algorithm first overclusters the space using $k$-means, and then applies spectral clustering to the cluster centers. This avoids the huge computational resource demand of standard spectral clustering.}

   \label{fig:manifold}
 \vspace{-2.5em}

\end{figure}

The intuition behind the OC-RA algorithm is as follows: in Figure~\ref{fig:manifold} we show, via a toy example, why the vanilla $k$-means algorithm generally performs poorly for identifying concepts in the latent space; it is because the sizes of concept clusters vary significantly. Therefore, to accurately find the concepts in the latent space, we first apply $k$-means to learn substantially more clusters than required, i.e., $K>C$. Then, we apply spectral clustering \cite{shi2000normalized} to the $K$ \textit{cluster centers} learned through $k$-means. The $K$ initial clusters are then reassigned to $C$ final clusters. Note that directly performing spectral clustering on the latent representations of the concept primitives could theoretically find clusters on the manifold. However, building a
similarity matrix and performing eigendecomposition for hundreds of thousands of concept primitives is computationally intractable, whereas the OC-RA algorithm circumvents this problem by limiting the scale of the spectral clustering to only the cluster centers from (overclustered) $k$-means. 
Overclustering has been used in previous works  \cite{caron2018deep,caron2019unsupervised,ji2019invariant} for other purposes: these algorithms apply overclustering as auxiliary optimization targets, whereas here we directly discover concepts from the latent space.

After this step, we have the concept/cluster labels for the primitives, which will be used to create pseudo segmentation labels in the concept refinement step.

\subsection{Concept Refinement via Neural Network Smoothing}
\label{sec:method:nn_smooth}

The clustering of primitives assigns each primitive $\mathbf{x}_i^{(j)}$ to a cluster label $y_i^{(j)}\in\mathcal{C}$. Reassembling all primitives of image $i$ with their cluster labels, we obtain pseudo segmentation labels $\mathbf{y}_ i\in \mathcal{C}^{H\times W}$. Specifically, for each pixel $\mathbf{x}_i[h,w]$ in the original image $\mathbf{x}_i$, the corresponding pseudo label is set to be the cluster label of the primitive that contains this pixel, i.e., if $\mathbf{x}_i[h,w]\in \mathbf{x}_i^{(j)}$, set $\mathbf{y}_ i[h,w] = y_i^{(j)}$. 

However, as a segmentation map, the pseudo label is far from perfect. For instance, if a part of a wall looks similar to part of the sky, a ``sky'' label could be assigned to a small piece of a wall. The concept refinement process is designed to rectify such problems by \textit{considering the global context of the patches}. Here, a segmentation network, composed of an encoder $f_\Phi$ and a decoder $g_\Theta$, is trained to predict the pseudo segmentation labels $\mathbf{y}_i$ for image $\mathbf{x}_i$. In previous steps, the scope of the encoder was limited to the local patch itself while encoding the images, and the context information was completely ignored. By exposing the neural network to full-image level information from the whole training set, it will learn a mapping from images to concept segmentation, which incorporates both local visual similarity \textit{and global contextual information}, following our Assumptions 2 and 3.

Due to the fact that the parameters for the encoder $\Phi$ will be updated, we denote the updated version as $\Phi'$ for disambiguation.
Then the loss function is defined as 
\begin{equation}
    l_{\text{seg}}(f_{\Phi'},g_{\Theta},\mathbf{x_i}) = - \sum_{h,w}\log \frac{\exp(g_\Theta(f_{\Phi'}(\mathbf{x_i}))_{h,w,\mathbf{y}_{i\lbrack h,w\rbrack}}}{\sum_{c=1}^{C}\exp(g_\Theta(f_{\Phi'}(\mathbf{x_i}))_{h,w,c}}.
\end{equation}
Since our initialization of the encoder $f_\Phi$ comes from a self-supervised network, it has an important quality for discovering concepts: invariance to common transformations (e.g., cropping, horizontal flipping, and change in saturation). Therefore, when training the segmentation network $f_{\Phi'}$ and $g_\Theta$, we also apply these transformations to the image and the pseudo labels.
In Section \ref{sec:experiments}, the ablation study shows that data augmentation is important for training the segmentation network.

\subsection{SegDiscover as a Latent Space Explanation Tool}
\label{sec:method:exp}

In addition to its main usage in discovering concepts, SegDiscover can be used as an explanation tool to visualize the latent space structure of a pretrained encoder $f_{\Phi}$. Since the discovered visual concepts are clusters in latent space, visualizing each concept separately, as in Figure \ref{fig:concepts}, may reveal what has been learned in the latent space of the network.
Furthermore, SegDiscover generates a separate segmentation map for each input image that specifies the location associated with each visual concept, which is form of local explanation, similar to saliency-based methods (but also with concept information). Thus, as an explanation tool, SegDiscover
combines the advantages of global concept-based explanation methods \cite{kim2018interpretability,zhou2018interpretable,ghorbani2019towards,yeh2020completeness} and local saliency-based explanation methods \cite{zeiler2014visualizing,simonyan2013deep,smilkov2017smoothgrad,selvaraju2017grad}. 

\section{Experiments}
\label{sec:experiments}
Our framework generates a segmentation map for each image in the dataset $\mathcal{X}$ with each segment associated with a discovered concept. In terms of evaluation, we would like to answer the following questions about the concepts discovered by the proposed method: 1. What concepts are discovered from the dataset? 2. Are many of the discovered concepts similar to real concepts in labeled datasets, despite the fact that they were obtained without labels? 3. How cohesive are the discovered concepts -- do most members of each concept represent the same type of entity? 4. What is the difference between the discovered concepts and concept maps if we use different encoders? 



The answers are provided as follows. In Section~\ref{sec:exp:setup}, we discuss the general experimental setup. In Section~\ref{sec:exp:concepts}, we visualize the concepts discovered to answer Question 1. In Section~\ref{sec:exp:segresults}, we evaluate the result as a semantic segmentation problem to answer Questions 2 and 3. Finally, in Section~\ref{sec:exp:nnexp}, we answer Question 4 and show how our method can be used as a neural network explanation tool.



\subsection{Experimental Setup}
\label{sec:exp:setup}
\paragraph{Dataset.} We performed experiments on the Cityscapes \cite{cordts2016cityscapes} and COCO-Stuff \cite{caesar2018coco} datasets; each image in these datasets contains multiple concepts in complex urban or natural scenes, and both datasets have rich annotations that (almost) cover the whole of each image, which we can use for evaluation (but not training because our method is fully unsupervised).
In line with \cite{ji2019invariant} and \cite{cho2021picie}, we merged the labels in the COCO-Stuff dataset into 27 categories (15 ``stuff'' categories and 12 ``things'' categories), and used the subset of images used by \cite{cho2021picie} in order to compare with their results. We also performed an analysis on the ADE20k dataset \cite{zhou2017scene}, with details and results in the \textit{supplementary materials}.

\paragraph{Implementation details.}
We used a ResNet-50 \cite{resnet} network as the backbone encoder $f_\Phi$, and four upsampling layers with residual connections to the end of each module in the backbone as the decoder $g_\Theta$, resulting in a Feature Pyramid Network \cite{lin2017fpn} structure when combined. Unless otherwise specified, the encoder $f_\Phi$ was always initialized by the DeepCluster-V2 \cite{caron2018deep,caron2020unsupervised} weights contrastively pretrained on ImageNet (without labels) \cite{Imagenet}.  

\subsection{Visualizing Discovered Concepts}
\label{sec:exp:concepts}

\begin{figure*}[t]
  \centering
   \includegraphics[width=0.9\linewidth]{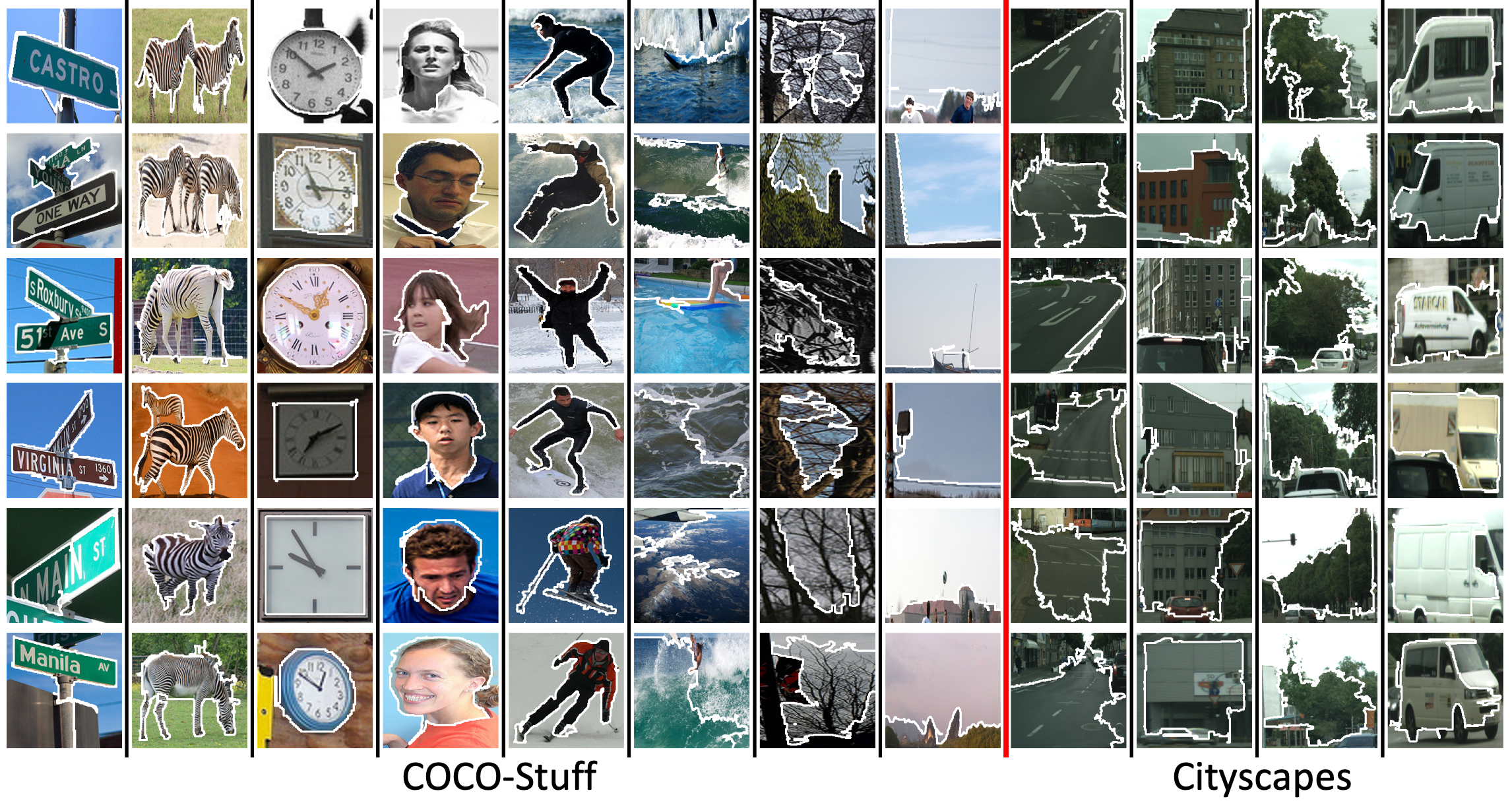}

   \caption{Samples of concepts discovered from the COCO-Stuff dataset (left 9 columns) and Cityscapes dataset (right 4 columns). Each column shows 6 concept primitives (with context) from a concept cluster, in which the primitive is highlighted by white boundaries. Notice the two different human clusters found in COCO-Stuff.
   More samples can be found in the \textit{supplementary materials}.}
   \label{fig:concepts}
\vspace{-0.5em}

\end{figure*}

Figure~\ref{fig:concepts} shows some of the concepts discovered by our method. We can see that the algorithm has successfully discovered meaningful concepts such as street signs, zebras, water and trees from the COCO Stuff dataset, as well as roads, buildings and cars from the Cityscapes dataset. Interestingly, SegDiscover found two different concepts related to humans on the COCO-Stuff dataset: human faces and skiers. Whereas the labels that come with the dataset indicate that the faces and skiers are both ``person,'' this misses the fact that these two types of images are very semantically different (face vs$.$ body). SegDiscover does not have the bias of the labelers and is able to distinguish between these two concepts. This simple example illustrates the power of SegDiscover's fully unsupervised approach for discovering concepts.

\subsection{Unsupervised Semantic Segmentation}
\label{sec:exp:segresults}
While our method is not designed for unsupervised semantic segmentation, it performs this task as a byproduct of the pipeline. Thus, we evaluate it with the metrics for unsupervised semantic segmentation. (We also evaluate its performance on explaining pretrained neural network latent spaces in Section~\ref{sec:exp:nnexp}.)

\paragraph{Baselines.}
Modified DeepCluster (MDC) \cite{caron2018deep,cho2021picie}
and PiCIE \cite{cho2021picie} are the current best unsupervised semantic segmentation algorithms for this task, and are thus used as baselines. We also compare to MaskContrast \cite{van2021unsupervised} which is the SOTA of another line of work on unsupervised semantic segmentation that focuses on segmentation of the foreground objects only.

\paragraph{Metrics.}
We evaluate the semantic segmentation performance of the proposed method under three settings: (1) completely unsupervised (``unsupervised''); (2) training a linear projection head on the network (``linear head''); (3) transfer learning with a linear projection head (``transfer'').

For the ``unsupervised'' setting, we perform a maximum matching proposed by \cite{van2021unsupervised} that simply matches the semantic group we identified to the class label that occupies the relative majority of pixels in the group. For fair comparison, the number of classes discovered by all methods is set to equal the number of classes of each dataset.
The segmentation results are then evaluated on three metrics: mean intersection over union (mIoU), weighted intersection over union (wIoU), and pixel accuracy (pAcc). We also qualitatively compare segmentation maps generated by several algorithms and the ground truth.\footnote{Some previous works in unsupervised semantic segmentation use the Hungarian algorithm to match the predicted semantic group to the ground truth labels. The match assignment is 1-to-1. However, this evaluation metric is flawed: it identifies that a semantic group matches a class label when pixels of that class do not even occupy a relative majority in this group.} For the ``linear head'' setting, we freeze the network weights and add a linear layer on the output of the network. The linear head is trained on the labeled dataset for 10 epochs. The ``transfer'' setting is similar to the ``linear head'' setting, but the linear head is trained and evaluated on a different dataset (e.g., the backbone network is trained on COCO-Stuff but we train a linear head on Cityscapes).

\paragraph{Results.}
Table \ref{tab:segresults} compares the segmentation performance of SegDiscover and the baselines on the COCO-Stuff and Cityscapes datasets. We evaluate SegDiscover with and without concept refinement, denoted in the table as ``SegDiscover$-$'' and ``SegDiscover.'' On both datasets, SegDiscover achieves the best segmentation performance on all three metrics (mIoU, wIoU and pAcc) under all three settings (unsupervised, linear head, and transfer), except mIoU of transfer learning on COCO-Stuff. Here, we have evidence that aiming for concept discovery actually enhanced our ability to perform segmentation: SegDiscover$-$, which has no concept refinement, is slightly worse than PiCIE on both datasets. Interestingly, the performance of SegDiscover under the transfer learning setting is better than that under the linear head setting on the Cityscapes dataset. This could be because COCO-Stuff is larger than Cityscapes and includes similar concepts.

\begin{figure}
    \begin{minipage}[t]{.463\textwidth}
      \centering
      \includegraphics[width=.99\linewidth,valign=t]{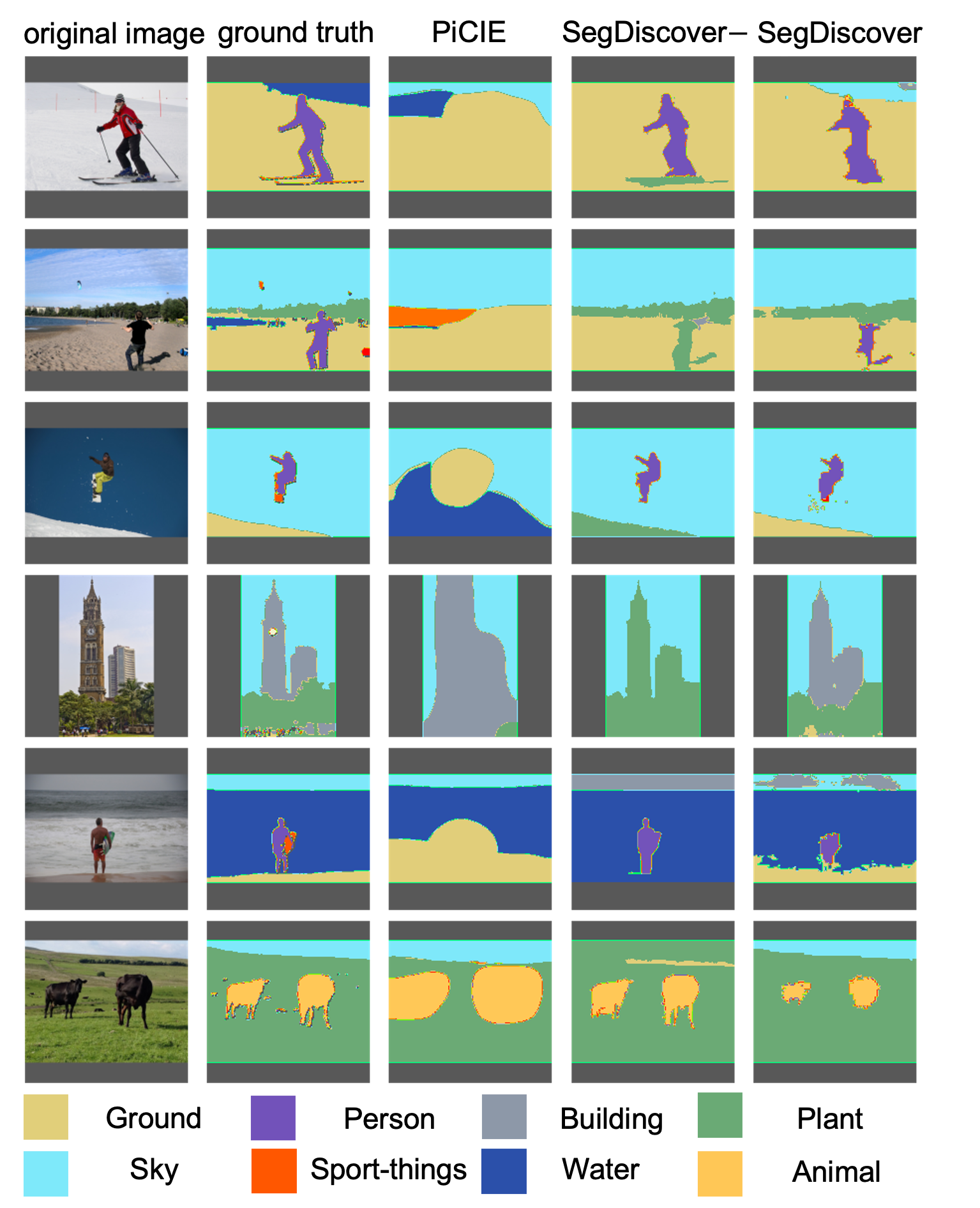}
      \caption{Unsupervised semantic segmentation maps generated on images from COCO-Stuff dataset. Colors in the segmentation are matched to the ground truth. Labels (provided in the dataset but not used for training) are provided in the legend. See \textit{supplementary materials} for more results.}
      \label{fig:qualitative_results}
    \end{minipage}%
    \hspace{0.1cm}
    \begin{minipage}[t]{.53\textwidth}
      \centering
      \includegraphics[width=.99\linewidth,valign=t]{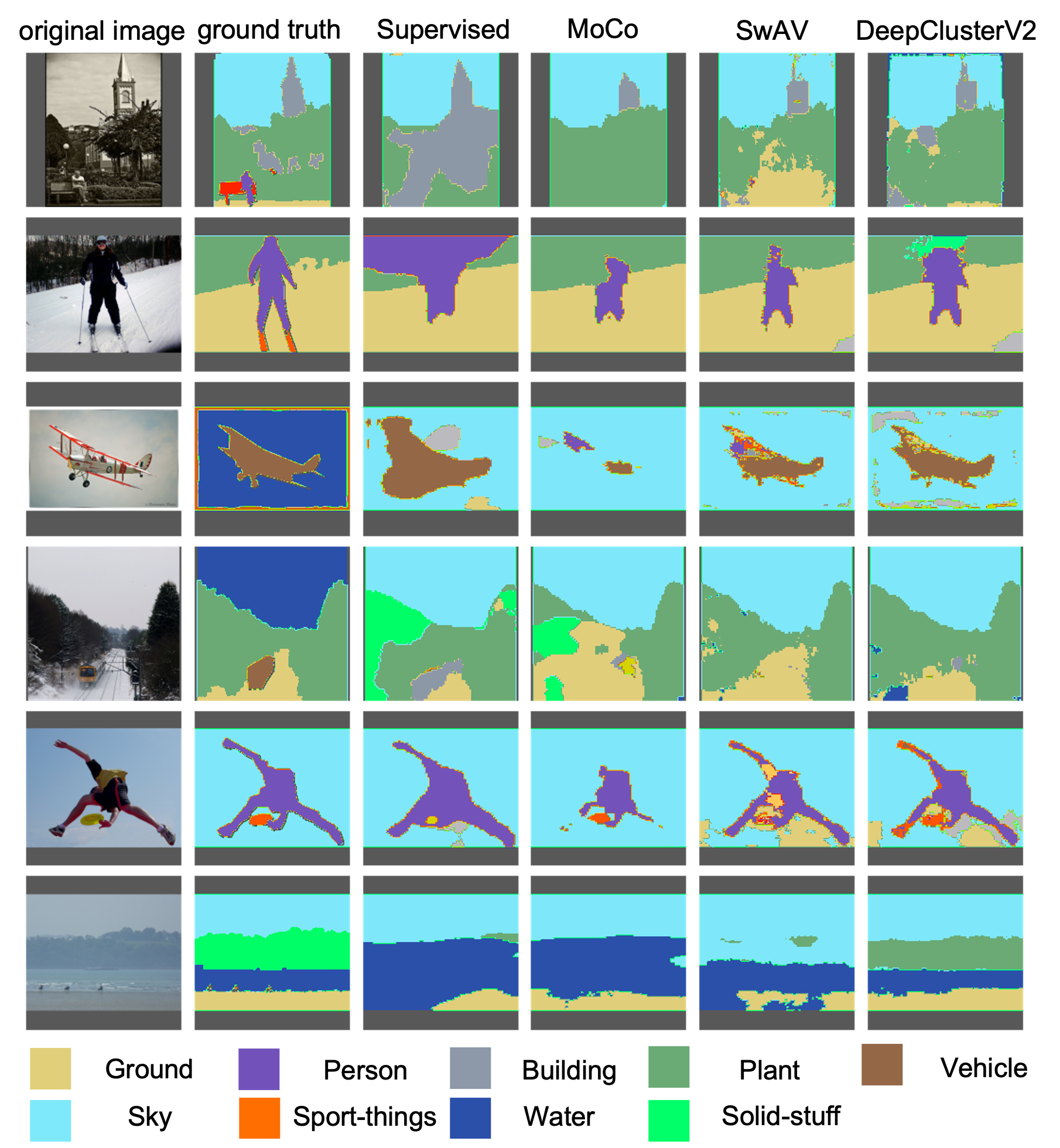}
      \caption{Unsupervised semantic segmentation maps generated by SegDiscover on COCO-Stuff images, using different pretrained encoders. Some errors in the annotations have been corrected by the algorithms (rows 3 and 4), which demonstrates the need for unsupervised algorithms. See \textit{supplementary materials} for more results.}
      \label{fig:compare}
    \end{minipage}
\end{figure}

Figure~\ref{fig:qualitative_results} shows examples of the segmentation maps generated by PiCIE \cite{cho2021picie}, SegDiscover$-$ and SegDiscover on the COCO-Stuff dataset. Comparing to PiCIE, the quality of the segmentation is generally better, because SegDiscover captures better boundaries from the image. While SegDiscover$-$ can create a good segmentation of the image, it sometimes assigns concepts incorrectly (e.g., SegDiscover$-$ believes the sky in row 5 is in the building concept, and generates segments containing multiple true concepts in rows 2 and 4). By incorporating information from the global context, and the whole dataset, SegDiscover's concept refinement corrects the error and improves the results.  


\begin{table}[]
\caption{Semantic segmentation results on COCO-Stuff and Cityscapes validation set. The methods are evaluated with three metrics (mIoU, wIoU, pACC) under three settings (unsupervised, linear head, transfer). In the transfer setting, results in the COCO-Stuff row are obtained when the concepts are discovered on Cityscapes with the linear head trained on COCO-Stuff, and vice versa. Note that SegDiscover$-$ does not train a segmentation network, so it is not applicable to the linear head and transfer setting.}
\begin{tabular}{c|l|ccc|ccc|ccc}
\hline
\multirow{2}{*}{Dataset} & \multicolumn{1}{c|}{\multirow{2}{*}{Method}} & \multicolumn{3}{c|}{Unsupervised} & \multicolumn{3}{c|}{Linear head} & \multicolumn{3}{c}{Transfer} \\ \cline{3-11} 
 & \multicolumn{1}{c|}{} & mIoU & wIoU & pAcc & mIoU & wIoU & pAcc & mIoU & wIoU & pAcc \\ \hline
 
\multirow{5}{*}{\textit{\begin{tabular}[c]{@{}c@{}}COCO-\\ Stuff\end{tabular}}} & MaskContrast & 8.86 & 7.80 & 23.03 & 24.54 & 42.88 & 59.33 & \textbf{18.59} & 37.52 & 54.43 \\
 & MDC & 5.68 & 19.22 & 34.12 & 5.91 & 19.97 & 36.65 & 10.56 & 31.81 & 50.64 \\
 & PiCIE & 11.79 & 32.70 & 50.67 & 12.62 & 32.98 & 50.06 & 6.00 & 20.73 & 37.62 \\
 & SegDiscover$-$ & 11.55 & 33.35 & 49.86 & -- & -- & -- & -- & -- & -- \\
 & SegDiscover & \textbf{14.34} & \textbf{40.11} & \textbf{56.53} & \textbf{35.29} & \textbf{58.75} & \textbf{72.89} & 15.10 & \textbf{38.26} & \textbf{55.76} \\ \hline
\multirow{5}{*}{\textit{Cityscapes}} & MaskContrast & 3.14 & 18.81 & 40.22 & 10.64 & 55.69 & 71.38 & 7.80 & 45.08 & 63.33 \\
 & MDC & 9.36 & 54.39 & 70.36 & 11.60 & 60.97 & 74.61 & 7.72 & 45.46 & 58.77 \\
 & PiCIE & 10.29 & 57.01 & 72.13 & 10.76 & 55.66 & 72.64 & 7.27 & 46.98 & 63.53 \\
 & SegDiscover$-$ & 10.65 & 55.00 & 70.73 & -- & -- & -- & -- & -- & -- \\
 & SegDiscover & \textbf{13.92} & \textbf{66.32} & \textbf{77.07} & \textbf{13.90} & \textbf{66.09} & \textbf{79.78} & \textbf{17.20} & \textbf{72.77} & \textbf{82.21} \\ \hline
\end{tabular}
 \label{tab:segresults}
\end{table}

\begin{table}[t]
  
  \begin{minipage}[t]{.55\linewidth}
  \caption{\textit{Ablation study 1}. Effect of different components in SegDiscover. Each row is one experiment where $\checkmark$ indicates the used component. Last row is complete SegDiscover.}
  \begin{tabular}{@{}lccc|ccc}
    \hline
    PM & OC-RA & NNS & NNS-DA & mIoU & wIoU & pAcc\\
    \hline
  & & & & 10.13 & 29.76 & 47.97 \\
\checkmark & & & & 9.97 & 29.81 & 46.07 \\
\checkmark & & \checkmark & & 10.89 & 32.69 & 49.42\\
\checkmark & \checkmark & & & 11.55 & 33.35 & 49.86\\
\checkmark & \checkmark & \checkmark & & 12.62 & 36.42 & 52.53 \\
\checkmark & \checkmark &  & \checkmark & \textbf{14.34} & \textbf{40.11}& \textbf{56.53} \\
    \hline
  \end{tabular}
  
  \label{tab:ablation1}
  \end{minipage}
  \hspace{0.1cm}
  \begin{minipage}[t]{.4\linewidth}
  \caption{\textit{Ablation study 2}. Effect of the parameter $K$ in the OC-RA algorithm on the COCO-Stuff dataset. }
    \begin{tabular}{l|c|cccl}
    \cline{1-5}
    Method & \textit{K} & mIoU & wIoU & pAcc &  \\ \cline{1-5}
    \multirow{4}{*}{SegDiscover} & 27 & 11.15 & 33.35 & 50.06 &  \\
     & 50 & 13.23 & 39.79 & 55.62 &  \\
     & 100 & \textbf{14.83} & {39.09} & \textbf{57.89} &  \\
     & 200 & 14.34 & \textbf{40.11} & 56.53 &  \\ \cline{1-5}
    PiCIE & n/a & 11.79 & 32.70 & 50.67 &  \\ \cline{1-5}
    \end{tabular}
    \label{tab:differentk}
    \end{minipage}
\end{table}




\paragraph{Ablation Studies}
Table \ref{tab:ablation1} shows an ablation study for SegDiscover on the COCO-Stuff dataset, where we measure the influence of various components of SegDiscover on segmentation performance. Specifically, we examine the following components of SegDiscover by replacing/removing combinations of them: the merging algorithm for primitives (PM), the overclustering+reassignment clustering algorithm (OC-RA), the concept refinement via neural network smoothing (NNS), and data augmentation for neural network smoothing (NNS-DA). Results in Table \ref{tab:ablation1} (comparing row 2 to row 4, and row 3 to row 5) show that the OC-RA clustering algorithm improves over plain k-means by 1-2 mIoU points, 3-4 wIoU points and 3-4 pAcc points. Neural network smoothing with data augmentation leads to a large (4 points) increase in wIoU and pAcc than simply performing neural network smoothing; our augmentation was designed to make the network invariant to common image transformations.

Table \ref{tab:differentk} shows the ablation study on the effects of the number of clusters $K$ in the overclustering step of the OC-RA algorithm. Note that no matter what $K$ is, we always output 27 classes, i.e., the ground truth classes. The results show the overclustering step can significantly improve the segmentation performance with good choices of $K$ (50, 100 and 200), with best performance at $K=\sim 100$.


\subsection{Segmentation as a NN Explanation Tool}
\label{sec:exp:nnexp}

\begin{table}[t]
  \centering
  \caption{Unsupervised semantic segmentation results of SegDiscover with different pretrained encoders on the COCO-Stuff dataset.}
  \begin{tabular}{@{}l|ccc@{}}
    \hline
    Pretrained encoder & mIoU & wIoU & pAcc\\
    \hline
    Supervised & 10.12 & 33.68 & 50.23\\
    MoCo &  13.38 & 36.41  & 54.62 \\
    SwAV & 13.62 & 35.30  & 54.65 \\
    DeepCluster-V2 & \textbf{14.34} & \textbf{40.11} & \textbf{56.53}\\
    \hline
  \end{tabular}
  \label{tab:segresults_pretrain}
\end{table}

As we discussed in Section~\ref{sec:method:exp}, by substituting the encoder $f_\Phi$ with a different pretrained model, we can obtain a new set of discovered concepts that potentially reveal the latent space structure of the network, and segmentation maps that highlight regions of interest for all discovered concepts in the images; highlighting regions of interest is a typical goal for an explanation method.

We compare the segmentation results of SegDiscover initialized with different self-supervised CNN encoders, including MoCo \cite{Mocov2}, SwAV \cite{caron2020unsupervised} and DeepCluster-V2 \cite{caron2020unsupervised,caron2018deep}, and SegDiscover initialized with a supervised CNN pre-trained on ImageNet\cite{Imagenet} for image classification. We denote SegDiscover intialized with these encoders as SegDiscover-MoCo, -SwAV, -DeepCluster, and -Supervised.

Table \ref{tab:segresults_pretrain} compares the mIoUs, wIoUs and pAccs of SegDiscover when different pretrained encoders are used. Interestingly, when the self-supervised encoders are used, SegDiscover performs better (by $>$4 pAcc points) than using the supervised encoder. It is possible that the supervised encoder learns too much bias in the ImageNet labels and thus transfer learning is less effective than if we had used a self-supervised encoder pre-trained on ImageNet. 
Among the self-supervised encoders, DeepCluster-V2 performs the best, while our method using \textit{any of the three self-supervised encoders outperforms the current state-of-the-art (PiCIE)}.



Figure \ref{fig:compare} shows the segmentation maps generated by SegDiscover with different pretrained encoders on COCO-Stuff images. To better demonstrate the segmentation results obtained with different pretrained encoders, here we directly match the overclustering results to the ground truth labels for visualization. 
Compared to initializing SegDiscover with self-supervised pretrained encoders, SegDiscover-Supervised tends to perform poorly in localizing objects (row 1-3), and can discover different types of concepts than initializing with self-supervised encoders (e.g., ``mountain'' in row 4).
We believe such differences result from the training procedure. When the encoder is trained for an image classification task, signals from human annotation will not only propagate to the \textit{object}, but also reach its \textit{context}. As a result, the segments containing salient objects tend to be bigger, containing part of their context. 

Among the self-supervised encoders, SegDiscover-MoCo tends to generate objects smaller than ground truth (rows 2, 3, 5). There are other differences in segmentation, for example, in row 1, SegDiscover-SwAV identifies parts of the building as  ``ground,'' and in row 6, SegDiscover-DeepClusterV2 identifies ``plants'' in the distant view; either labeling could arguably be ground truth.

\section{Discussion}
Unsupervised concept discovery is a challenging, well-known problem in interpretable machine learning \cite{rudin2021interpretable}. There are very few works that have attempted this challenging task for natural images.
SegDiscover combines ideas from semantic segmentation and concept-based explainable machine learning. It not only achieves state-of-the-art unsupervised segmentation performance but also discovers cohesive and meaningful visual concepts from images without any supervision. In addition, it can be used as an explanation tool to visualize the latent space of a pre-trained network: it provides \textit{global} concept-based explanations, but also maps them back to \textit{local} images like saliency-based explanations.

There is much room for future work. For example, for simplicity, the proposed framework assumes there exists a dense segmentation for an image into meaning visual concepts. However, an image might contain noise or other pixels that do not necessarily belong to any meaningful visual concept. Moreover, although the proposed framework can reduce biases from the annotation
in the discovered concepts as the entire process is unsupervised, biases in the data collection process may still exist. If certain concepts are underrepresented in the dataset, they may fail to be recognized as visual concepts by our framework. SegDiscover lays the foundation for solving these problems, and designing fully-interpretable neural networks for real world computer vision applications.

\clearpage
%
%
\bibliographystyle{splncs04}
\bibliography{segdiscover}
\end{document}


\pagestyle{headings}
\mainmatter
\def\ECCVSubNumber{5365}  

\title{SegDiscover: Visual Concept Discovery via Unsupervised Semantic Segmentation -- Supplementary Material} 

\titlerunning{SegDiscover}
%

\author{Haiyang Huang\thanks{equal contribution}
 \and
Zhi Chen\printfnsymbol{1} \and
Cynthia Rudin}

\authorrunning{H. Huang et al.}
%
\institute{Duke University, Durham, NC 27708, USA\\
\email{\{hyhuang, zhichen, cynthia\}@cs.duke.edu}}
\maketitle


\section{More Implementation Details}
\subsection{Primitive Segmentation}
For the initial concept primitive segmentation, we use the scikit-image \cite{scikit-image} implementation of the Felzenszwalb algorithm \cite{felzenszwalb2004efficient}. We choose the parameter $scale=1000$ for all our experiments. Due to difference in figure size and contrast, we choose the gaussian kernel size $\sigma = 0.3$ on the COCO-Stuff dataset, and $\sigma=0.1$ on the Cityscapes and ADE20k dataset. 

To accommodate to different image sizes, the hyperparameter $min\_size$ is dynamically defined. Specifically, we use the following strategy to define $min\_size$:
\begin{equation}
    min\_size[i] = \max\left(\frac{H[i]}{768} \times \frac{W[i]}{1024} \times 5000, 250\right) 
\end{equation}
in which $H[i]$ and $W[i]$ are the height and width of image $i$, respectively. Such strategy ensures robustness of primitive segmentation with respect to image size. 

\subsection{Primitive Merging}
To further improve the quality of concept primitives and reduce irregular patches, we merge some of the primitives. See Algorithm~\ref{alg:pm} for the pseudocode of the Primitive Merging algorithm.
\begin{algorithm}
    \caption{Pseudocode of Primitive Merging}
    \label{alg:pm}
    \begin{algorithmic}

    \REQUIRE
    \qquad\\ 

            $\bullet$ $\mathbf{x}_i$ -- an image.
            
            $\bullet$ $\mathbf{x}_i^{(j)}$, $j=1, \ldots, n$ -- the set of concept primitives found by superpixel algorithms in the image.

    \ENSURE \quad


    \FOR{$j=1$ \textbf{to} $n$}
        \STATE{
            $\bullet$ construct adjacency list for the patch $j$ of image $i$, i.e. $\mathbf{x}_i^{(j)}$ by computing a binary dilation with connectivity 2. 
            In other words, two concept primitives are neighbors if they each contain at least one pixel of a single 2 $\times$ 2 square.
            
            $\bullet$ sort the adjacency list for patch $j$ of image $i$, i.e. $\mathbf{x}_i^{(j)}$, by the number of neighboring pixels of other patches $\mathbf{x}_i^{(k)}$ that are shared with $\mathbf{x}_i^{(j)}$. Keep the top 3 neighbor concept primitives only.

            $\bullet$ compute the average hue within the patch $\mathbf{c}_i^{(j)}$.
            
            $\bullet$ compute the perimeter-to-area ratio 
            
            $p_i^{(j)} = \frac{Perimeter_i^{(j)}}{\sqrt{Area_i^{(j)}}}$
            
            $\bullet$ initialize the merge indicator $m_i^{(j)} = 0$.
        }
    \ENDFOR
    \FOR{$j=1$ \textbf{to} $n$}
        \FOR{$k=1$ \textbf{to} $3$}
        \STATE{
            $\bullet$ denote the $k$-th top adjacent concept primitive of patch $\mathbf{x}_i^{(j)}$ as $\mathbf{x}_i^{(j,k)}$.
            
            $\bullet$ If all the following requirements satisfy, merge $\mathbf{x}_i^{(j)}$ with $\mathbf{x}_i^{(j,k)}$, set $m_i^{(j)} = 1$ and \textbf{break}.
            
            $\cdot$ $\mathbf{c}_i^{(j)} - \mathbf{c}_i^{(j,k)} < 40$ \;\;(similar avg hue)
            
            $\cdot$ $m_i^{(j)} = 0$ \;\;(not already merged)

            $\cdot$ $Area_i^{(j)} < 0.001 \times Area_i \times (p_i^{(j)})^2$ or $p_i^{(j)} > 9$ (small area relative to perimeter)
        }
        \ENDFOR
    \ENDFOR

    \STATE \textbf{return} $\mathbf{x}_i^{(j)}$, $j=1, \ldots m$ -- the merged concept primitives
    \end{algorithmic}
\end{algorithm}

\subsection{Concept Discovery}
The concept clusters are obtained via mini-batch k-means using the scikit-learn implementation \cite{sklearn_api}. We use $\ell_2$ distance as the distance metric, and apply a batch size of 1000 and a maximum iteration of 10000, with an early stopping strategy with threshold of 100. The number of clusters $k$ is set to be the number of known classes in each dataset from the ground truth labels. Coincidentally, $k=27$ for both the COCO-Stuff and Cityscapes dataset, and $k=150$ for ADE20k. 

For overclustering, we set $k^*=200$ for COCO-Stuff, and $k^*=400$ for ADE20k.  Spectral clustering with kernel $\sigma=1\times 10^{-5}$ is applied on the cluster centers to further merge the $k^*$ clusters into $k$ clusters. We did not apply overclustering for Cityscapes. 
\subsection{Concept Refinement}
We trained the encoder $f_\Phi$ and decoder $g_\Theta$ together for 30 epochs on the COCO-Stuff dataset, and 50 epochs on the Cityscapes dataset. We use an SGD-with-momentum-optimizer with learning rate $\eta=1\times10^{-3}$, momentum 0.9 and weight decay $5\times10^{-4}$. 

\subsection{Unsupervised Semantic Segmentation Experiments}
For PiCIE \cite{cho2021picie}, we directly utilized the weight and the implementation provided by the authors. For MDC \cite{caron2018deep,cho2021picie}, since the authors did not release the weights, we trained the algorithm using the provided implementation with default hyperparameters provided in the table. The obtained segmentation was then interpolated to the size of the original image for evaluation. For MaskContrast, we pretrained the models using supervised saliency masks, and directly utilized the k-means clustering provided by the authors.

\section{More Quantitative Segmentation Results}
Table \ref{tab:segresults_hungarian} shows the mIoUs and pAccs evaluated when the clusters are assigned using the Hungarian Algorithm. SegDiscover consistently overperforms the baselines except on the pixel accuracy metric on the Cityscapes dataset. This is caused by the fact that the evaluation metric becomes meaningless in the face of large class imbalances: a large portion of pixels in the Cityscapes dataset are from roads and buildings, and SegDiscover will create multiple different concept clusters to represent these most common categories (see Figure~\ref{fig:app_concs}). 
The Hungarian Algorithm will create a bijection between concept clusters and categories, hence only one of the clusters can be assigned to the correct categories. It is even possible for this algorithm to assign a class label to a cluster that has 
\textit{no} elements of the class in the cluster, which renders the evaluation meaningless.
Notice that the result for PiCIE is slightly different from the results reported in its original paper, because the original results appear to have been incorrectly evaluated on centercropped images.

\begin{table}
  \centering
  \caption{Results on unsupervised semantic segmentation of COCO-Stuff and Cityscapes validation set. Clusters are assigned using Hungarian matching.}
  \begin{tabular}{@{}l|c|ccc@{}}
    \hline
    Method & Dataset  & mIoU & wIoU & pAcc\\
    \hline
    MDC & \multirow{5}{*}{\textit{COCO-Stuff}} & 5.96 & 19.22 & 28.41 \\
    MaskContrast &  & 5.75 & 5.90 & 20.25 \\
    PiCIE &  & 12.54 & 31.96 & 46.03 \\
    SegDiscover$-$ & & 12.48 & 32.70 & 42.96\\
    SegDiscover & & \textbf{14.20} & \textbf{37.77} & \textbf{49.18}\\
    \hline
    
    MDC & \multirow{5}{*}{\textit{Cityscapes}}  & 6.10 & 20.78 & 23.96 \\
    MaskContrast &  & 3.96 & 16.54 & 35.49 \\
    PiCIE & & 10.15 & \textbf{47.77} & \textbf{59.88}\\
    SegDiscover$-$ & & 7.56 & 26.55 & 30.77\\
    SegDiscover & & \textbf{10.29} & 38.07 & {42.99}\\
    \hline
  \end{tabular}
  \label{tab:segresults_hungarian}
\end{table}

\section{More Concept Discovery Results}
\begin{figure*}[t]
  \centering
   \includegraphics[width=0.95\linewidth]{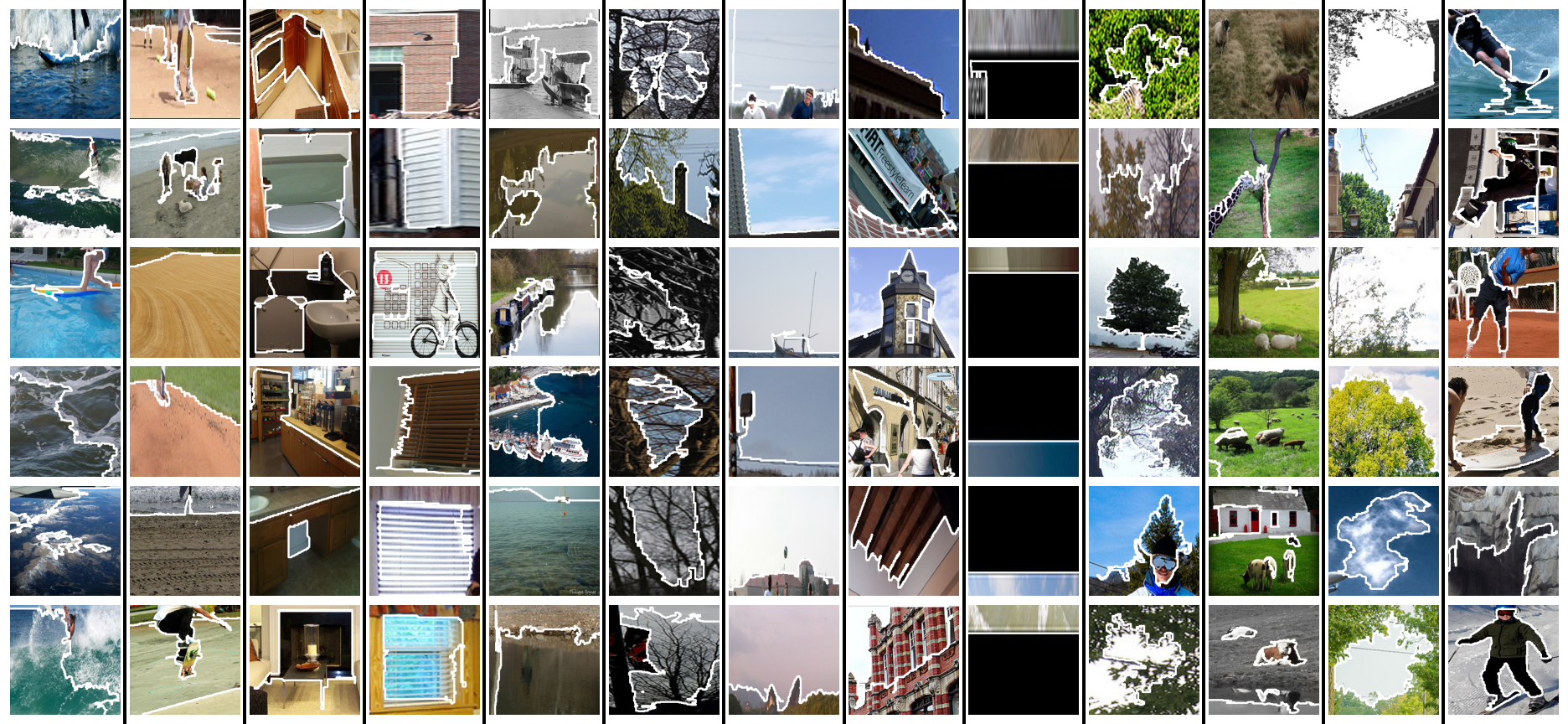}

   \caption{Samples of concepts discovered from the COCO-Stuff dataset \cite{caesar2018coco} (part 1 -- second part is in the next figure). Each column shows 5 concept primitives (with context) from a concept cluster, in which the primitive is highlighted by white boundaries.}
   \label{fig:app_concoco1}
\end{figure*}

\begin{figure*}[t]
  \centering
   \includegraphics[width=0.95\linewidth]{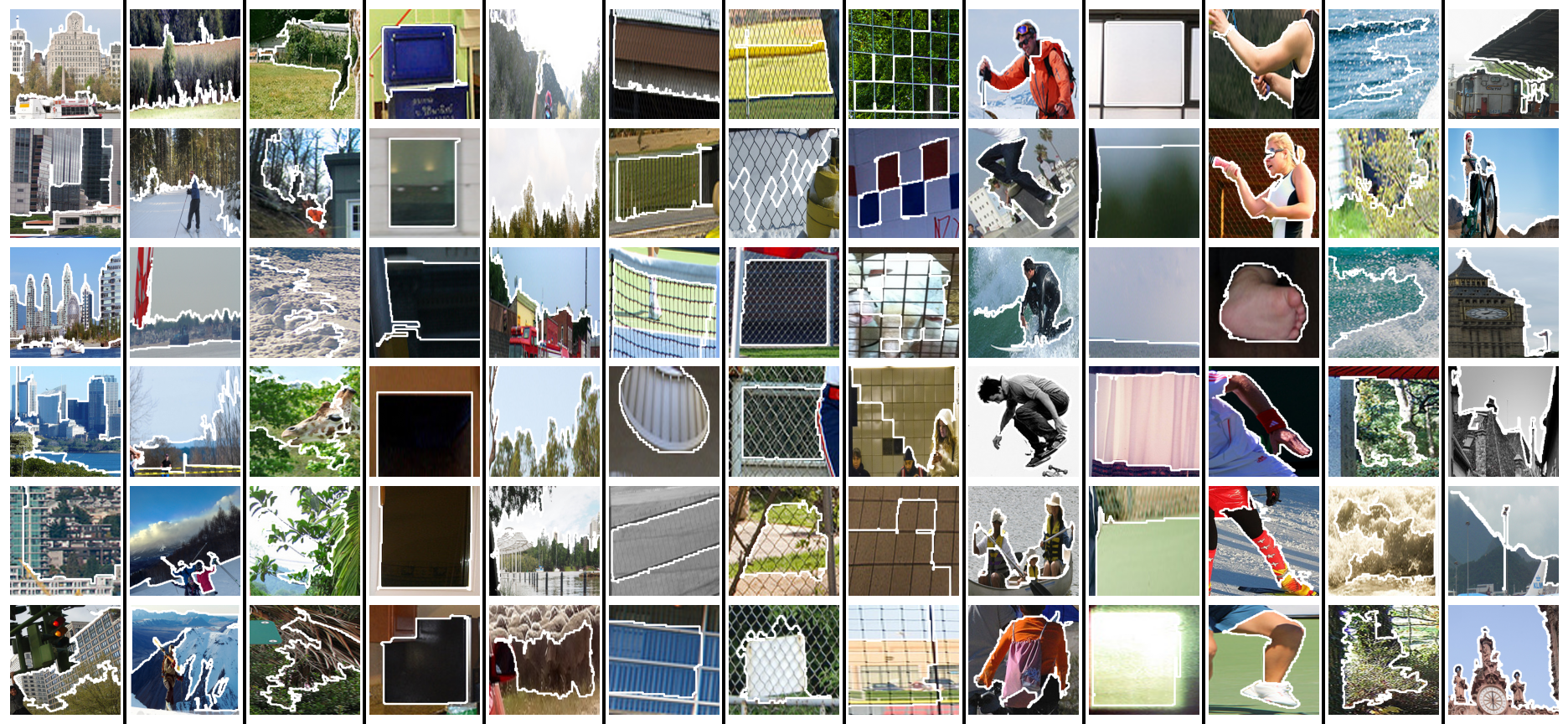}

   \caption{More samples of concepts discovered from the COCO-Stuff dataset \cite{caesar2018coco}  (part 2).}
   \label{fig:app_concoco2}
\end{figure*}

\begin{figure*}[t]
  \centering
   \includegraphics[width=0.95\linewidth]{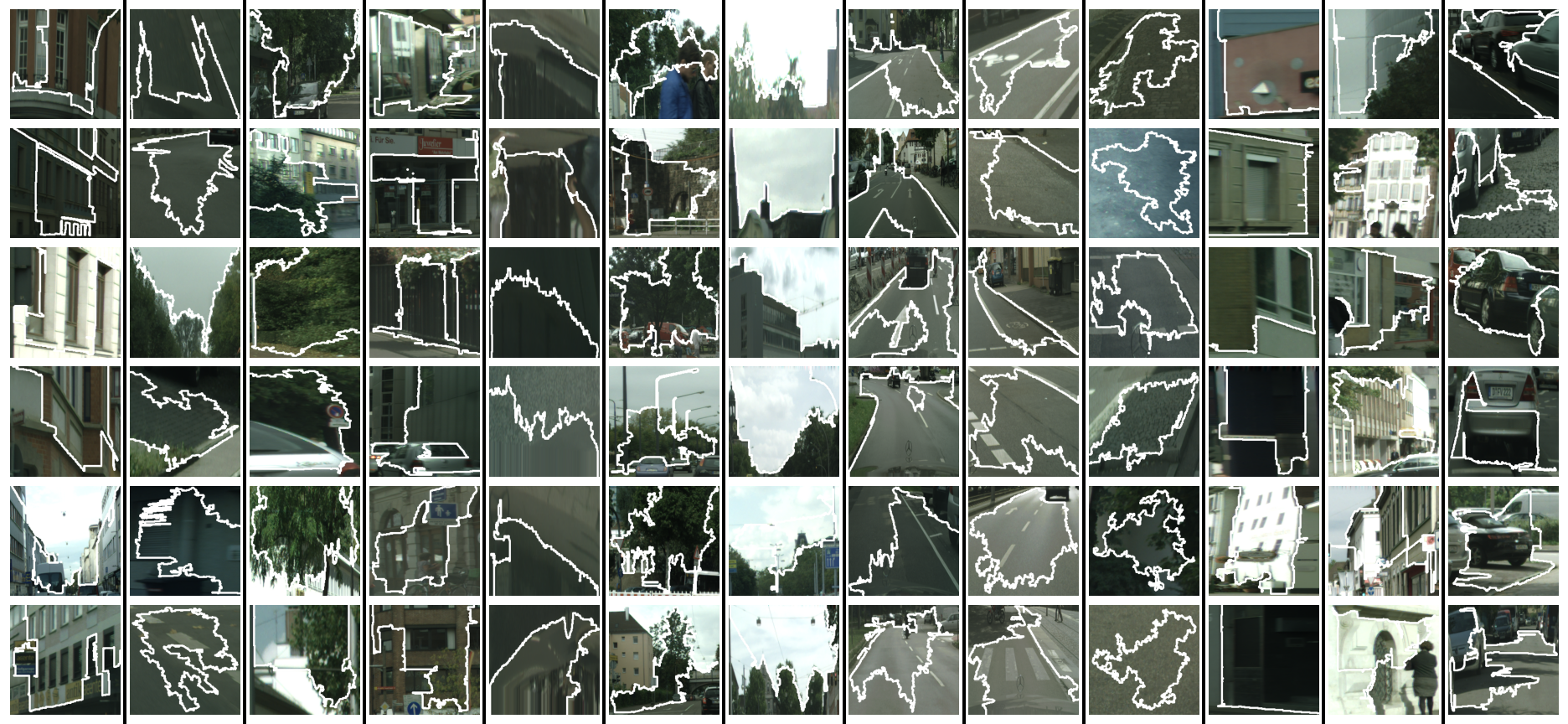}

   \caption{Samples of concepts discovered from the Cityscapes dataset \cite{cordts2016cityscapes}.}
   \label{fig:app_concs}
\end{figure*}

\begin{figure*}[t]
  \centering
   \includegraphics[width=0.95\linewidth]{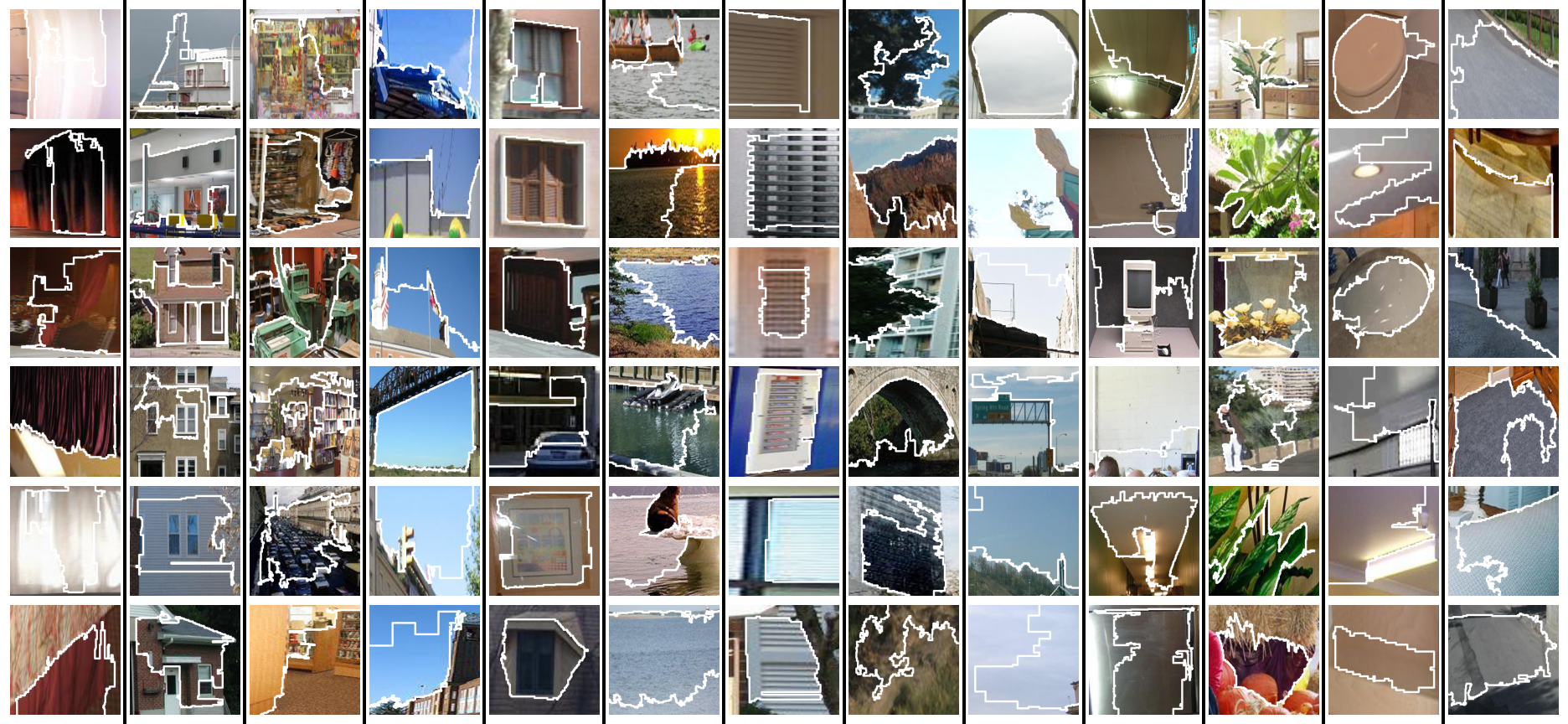}
   \caption{Samples of concepts discovered from the ADE20k dataset \cite{zhou2017scene}  (part 1).}
   \label{fig:app_conade1}
\end{figure*}

\begin{figure*}[t]
  \centering
   \includegraphics[width=0.95\linewidth]{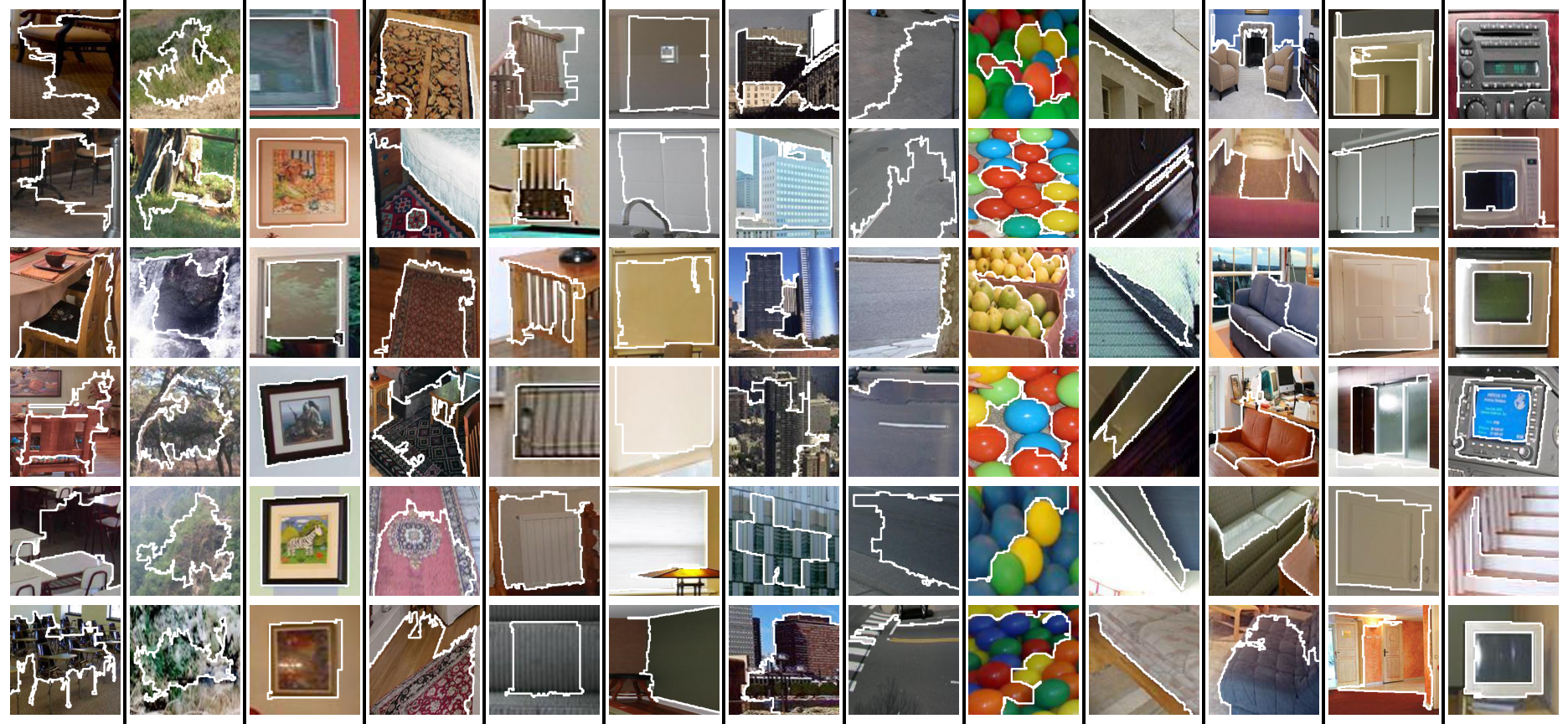}
   \caption{More samples of concepts discovered from the ADE20k dataset \cite{zhou2017scene}  (part 2).}
   \label{fig:app_conade2}
\end{figure*}

In this section, we provide more visualization results for concepts discovered by SegDiscover on different datasets. 
Figures~\ref{fig:app_concoco1} and~\ref{fig:app_concoco2} show more concepts obtained from the COCO-Stuff \cite{caesar2018coco} dataset. 
Figure~\ref{fig:app_concs} shows more concepts obtained from the Cityscapes \cite{cordts2016cityscapes} dataset. 
Figures~\ref{fig:app_conade1} and~\ref{fig:app_conade2} show some concepts obtained from the ADE20k \cite{zhou2017scene} dataset.

\section{More Unsupervised Semantic Segmentation Results}
\begin{figure*}[t]
  \centering
   \includegraphics[width=0.7\linewidth]{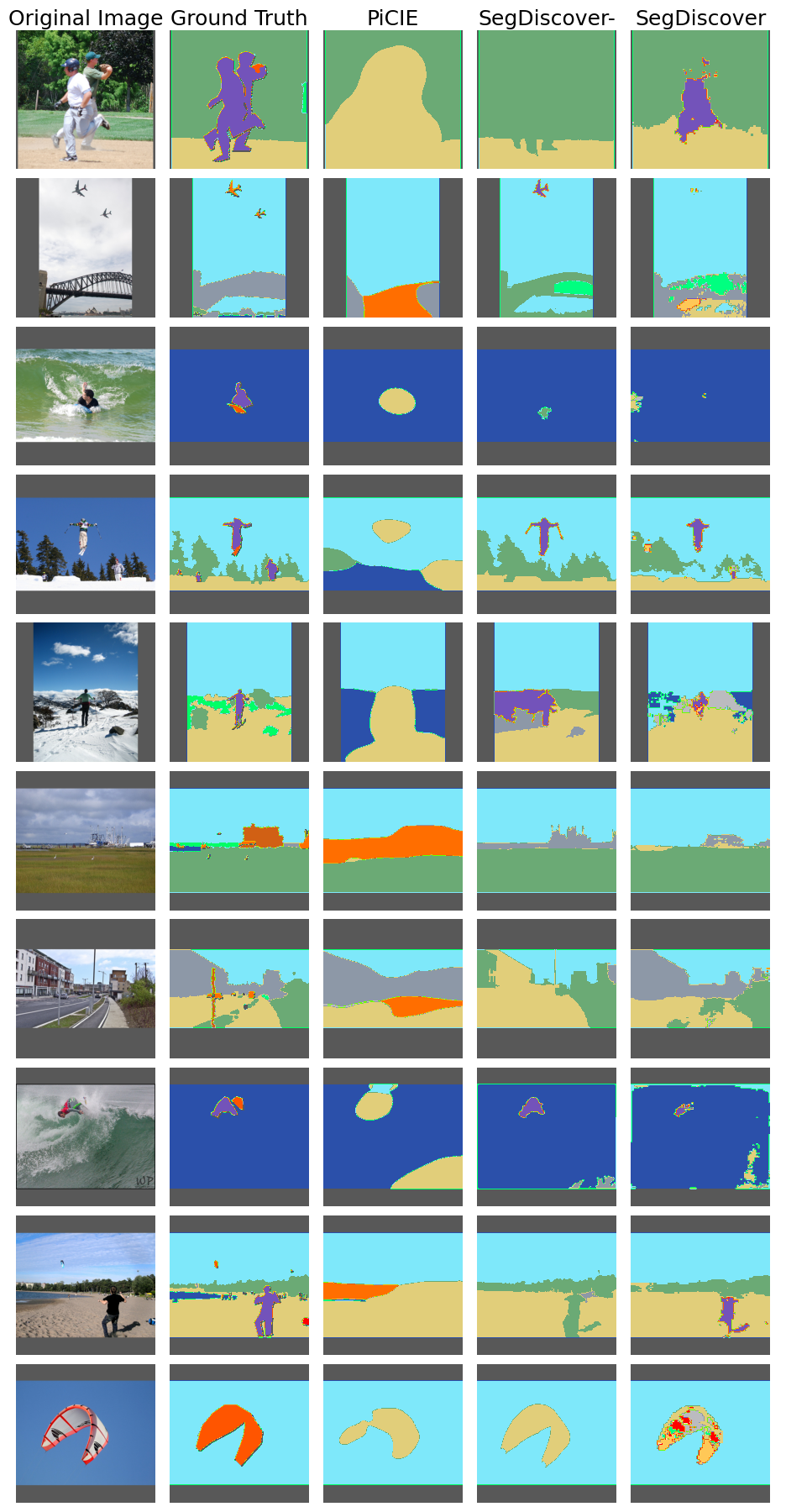}
   \caption{Samples of unsupervised semantic segmentation generated on images from COCO-Stuff dataset  (part 1). Colors in the segmentation are matched to the ground truth. The column heading shows the methods being used for segmantic segmentation.}
   \label{fig:app_segcoco1}
\end{figure*}

\begin{figure*}[t]
  \centering
   \includegraphics[width=0.7\linewidth]{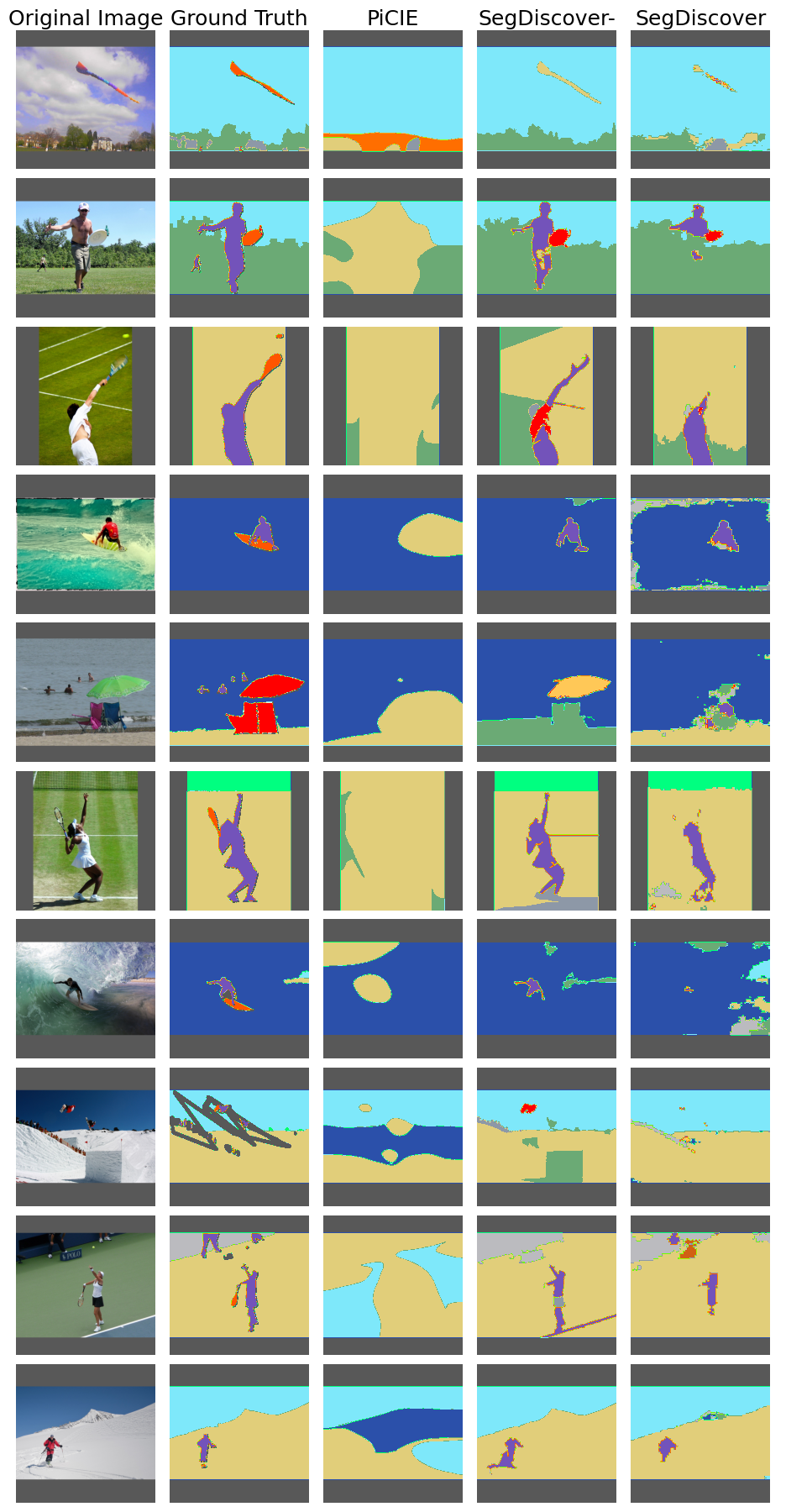}
   \caption{More samples of unsupervised semantic segmentation generated on images from COCO-Stuff dataset  (part 2). The column heading shows the methods being used for segmantic segmentation.}
   \label{fig:app_segcoco2}
\end{figure*}

\begin{figure*}[t]
  \centering
   \includegraphics[width=0.7\linewidth]{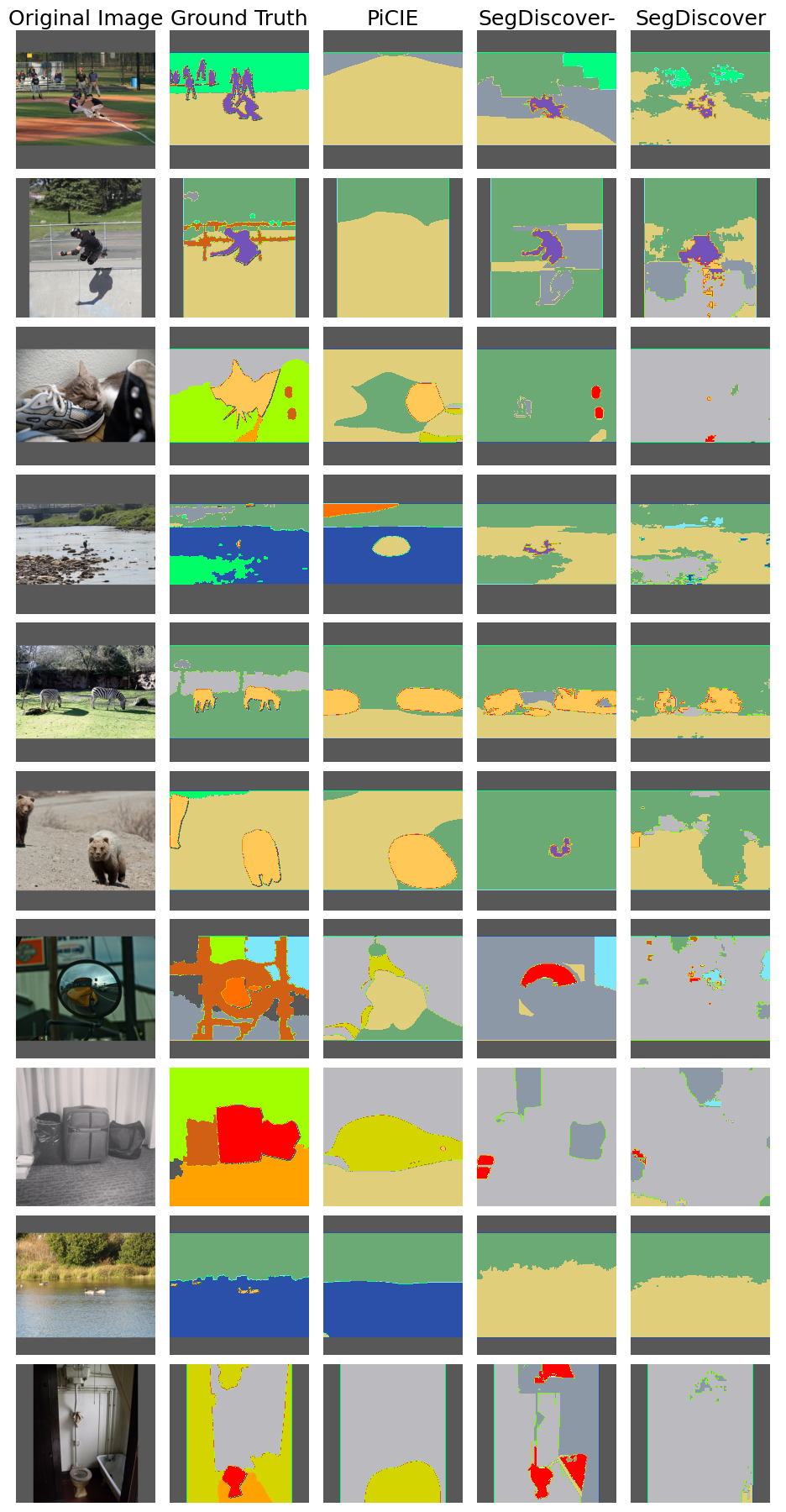}
   \caption{More samples of unsupervised semantic segmentation generated on images from COCO-Stuff dataset  (part 3). The column heading shows the methods being used for segmantic segmentation. This figure shows cases where SegDiscover perform poorly.}
   \label{fig:app_segcoco3}
\end{figure*}

\begin{figure*}[t]
  \centering
   \includegraphics[width=0.7\linewidth]{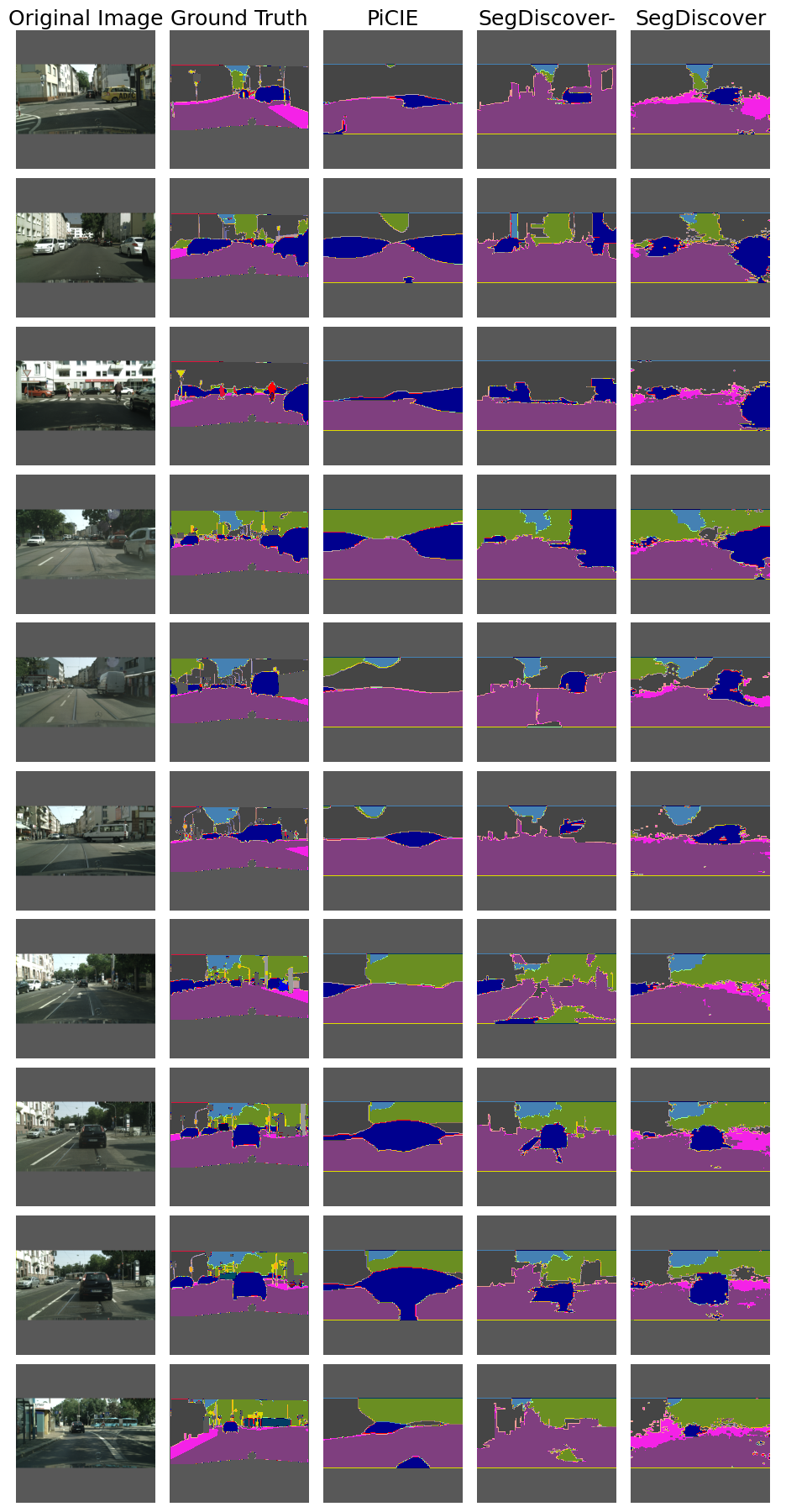}
   \caption{Samples of unsupervised semantic segmentation generated on images from Cityscapes dataset (part 1). Colors in the segmentation are matched to the ground truth. The column heading shows the methods being used for segmantic segmentation.}
   \label{fig:app_segcs1}
\end{figure*}

\begin{figure*}[t]
  \centering
   \includegraphics[width=0.7\linewidth]{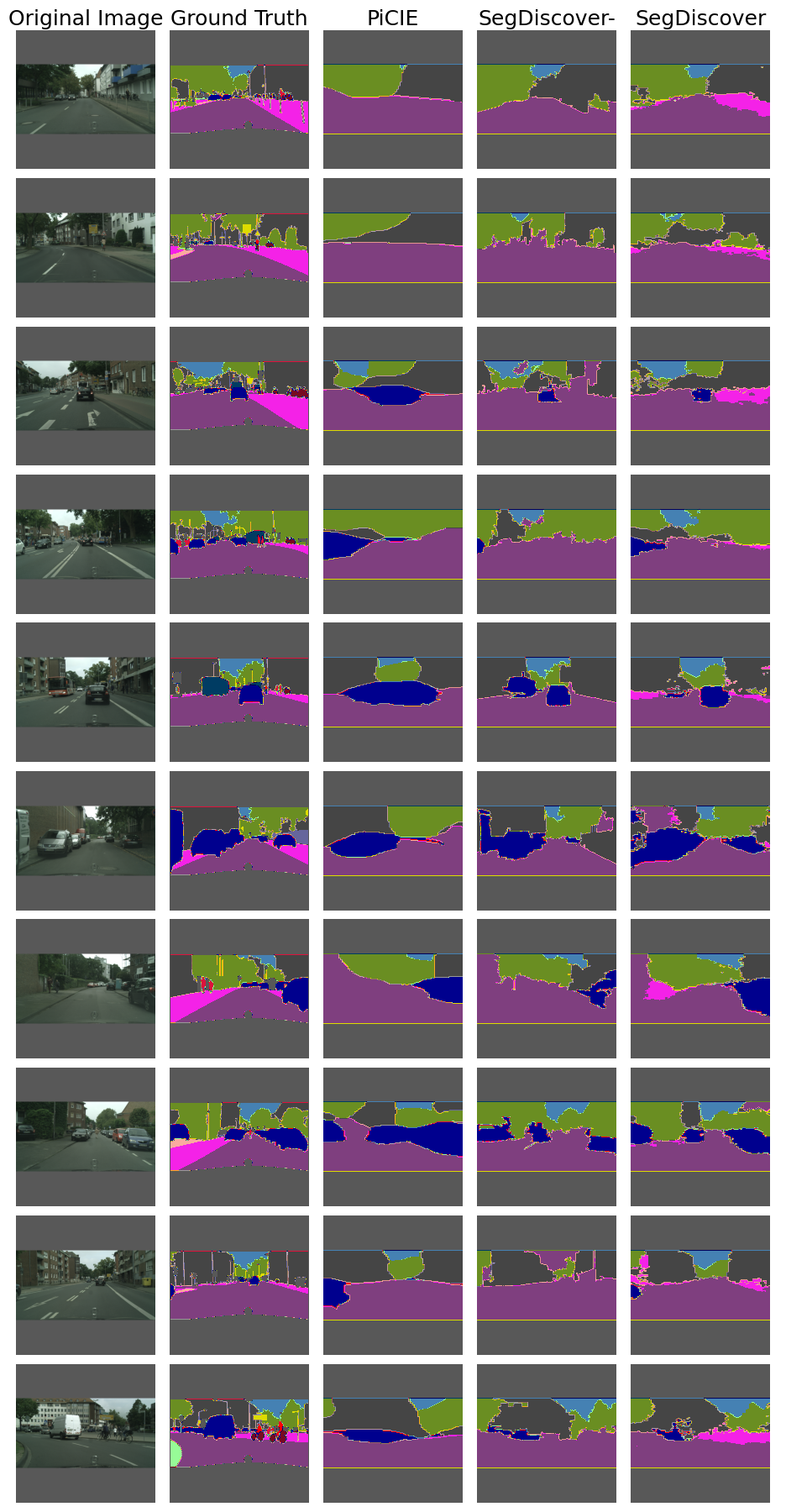}
   \caption{More samples of unsupervised semantic segmentation generated on images from Cityscapes dataset  (part 2). The column heading shows the methods being used for segmantic segmentation.}
   \label{fig:app_segcs2}
\end{figure*}

\begin{figure*}[t]
  \centering
   \includegraphics[width=0.7\linewidth]{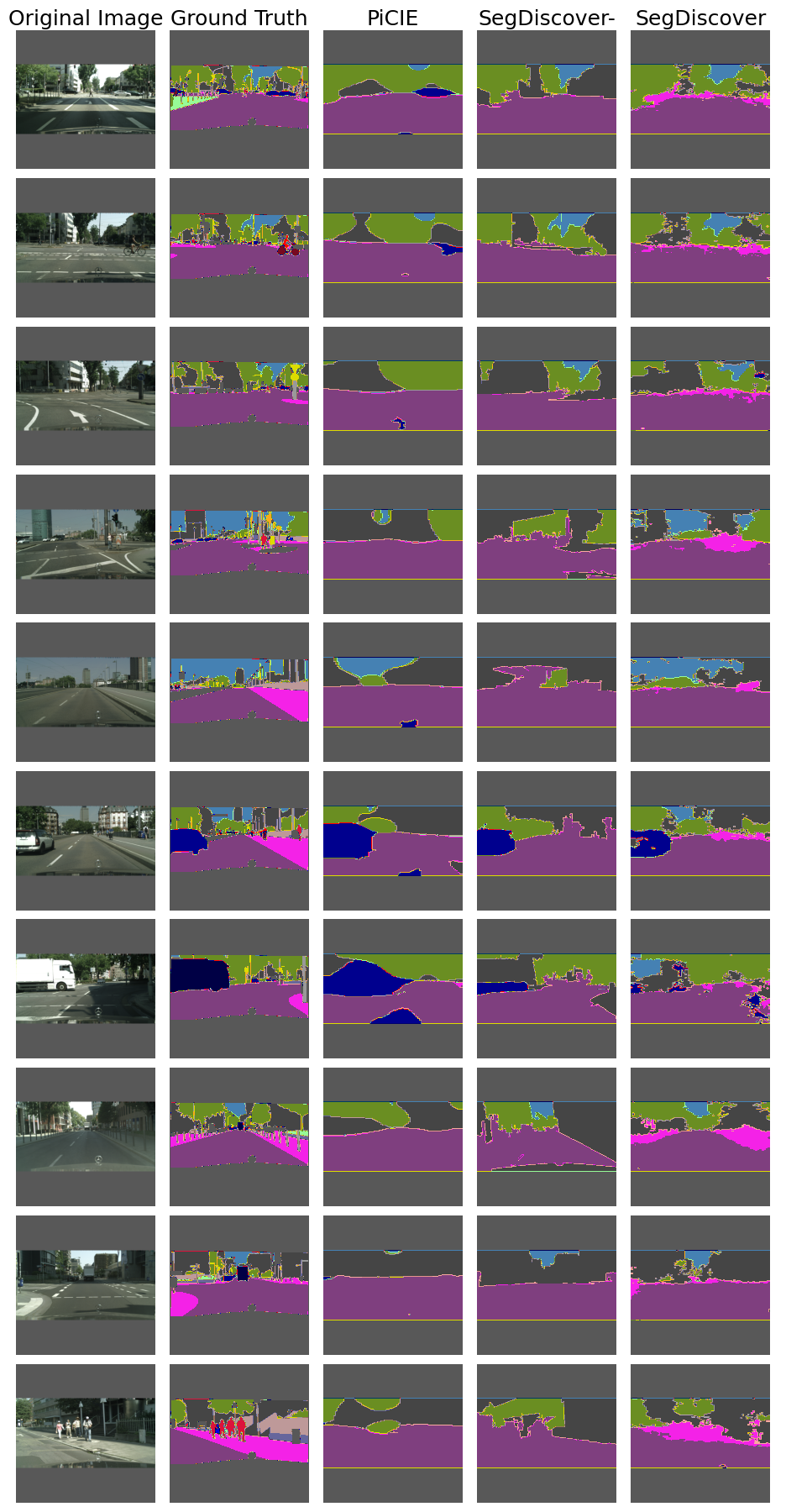}
   \caption{More samples of unsupervised semantic segmentation generated on images from Cityscapes dataset  (part 3). The column heading shows the methods being used for segmantic segmentation.}
   \label{fig:app_segcs3}
\end{figure*}

In this section, we provide more unsupervised semantic segmentation results. 
Figures~\ref{fig:app_segcoco1}, ~\ref{fig:app_segcoco2} and~\ref{fig:app_segcoco3} show more segmentation results on the COCO-Stuff \cite{caesar2018coco} dataset. Notice that Figure~\ref{fig:app_segcoco3} shows cases where SegDiscover performs poorly. This illustrates that SegDiscover may fail on distant objects (such as row 1 and 5) or objects with vague boundaries (such as row 3 and 6).
Figures~\ref{fig:app_segcs1}, ~\ref{fig:app_segcs2}, and~\ref{fig:app_segcs3} show more segmentation results obtained from the Cityscapes \cite{cordts2016cityscapes} dataset.


\section{More Results on Neural Network Explanation}
Table~\ref{tab:discoveredclasses} shows the classes in the COCO-Stuff dataset that were discovered by SegDiscover with different initializations. Here we define a class to be ``discovered'' if there is a cluster assigned to the class through maximum assignment. The supervised pretrained and SwAV initialization perform well according to this metric.

\begin{figure*}[t]
  \centering
   \includegraphics[width=0.7\linewidth]{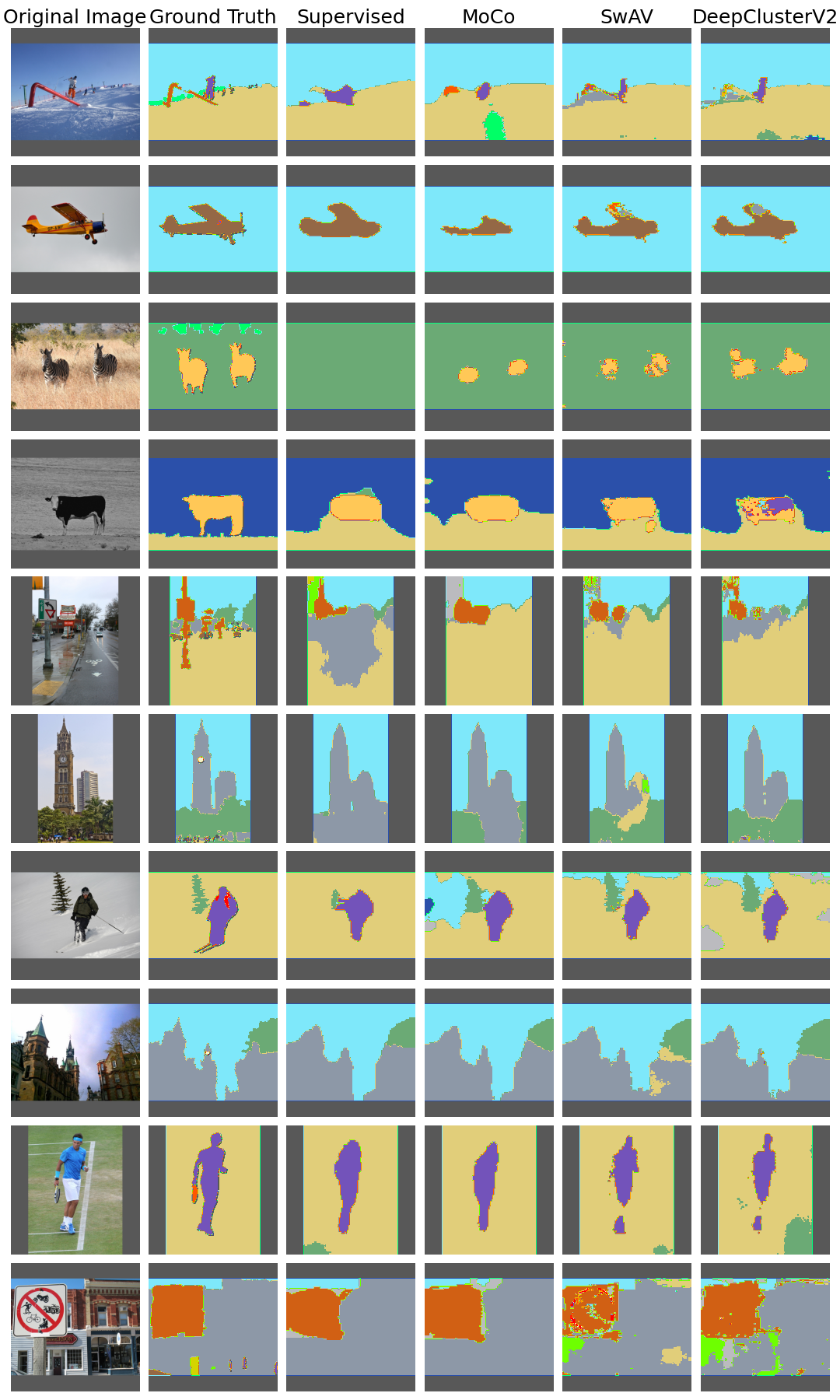}
   \caption{Samples of unsupervised semantic segmentation generated by SegDiscover with different initializations on images from COCO-Stuff dataset  (part 1). The column heading shows the initialization being used by SegDiscover. Colors in the segmentation are matched to the ground truth.}
   \label{fig:app_segcom1}
\end{figure*}

\begin{figure*}[t]
  \centering
   \includegraphics[width=0.7\linewidth]{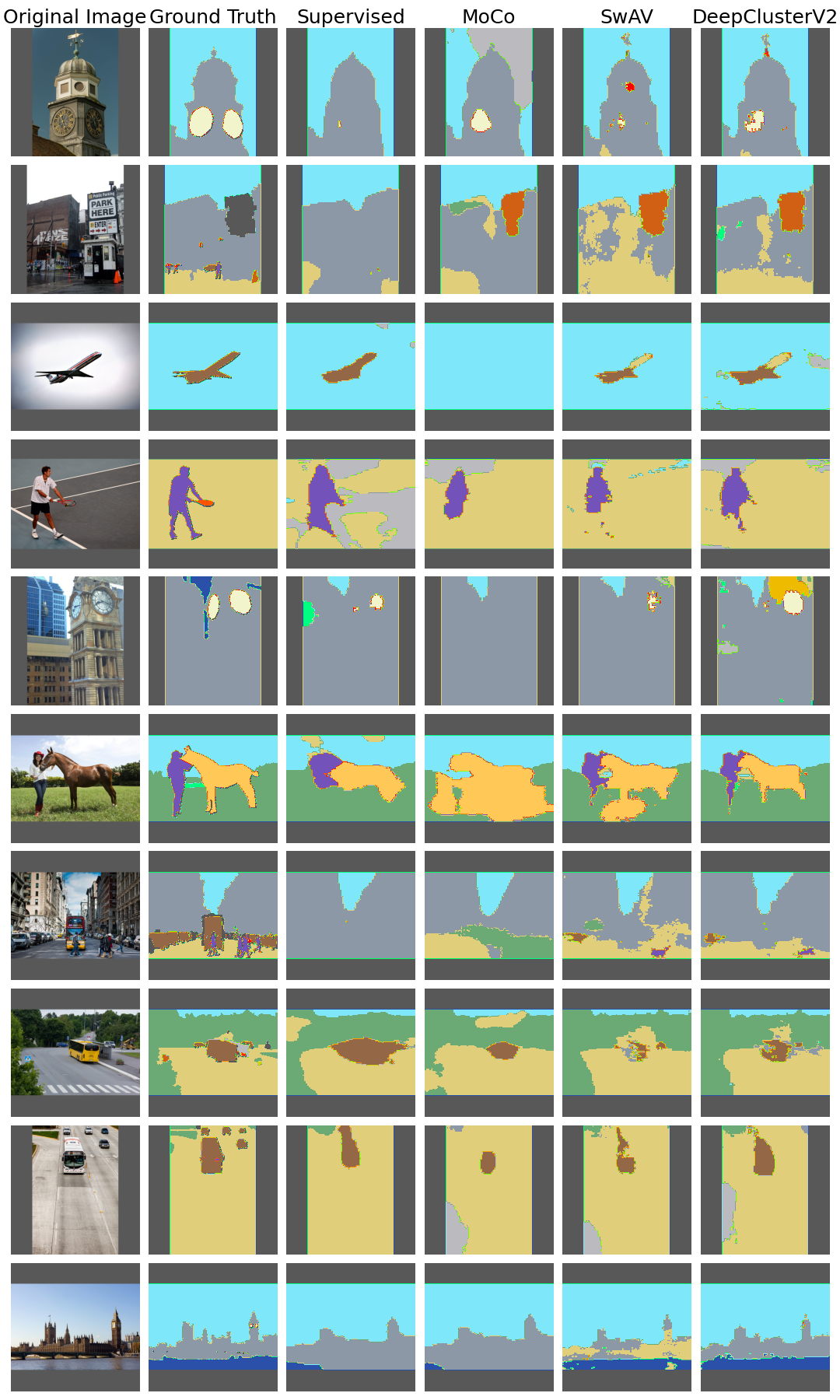}
   \caption{More samples of unsupervised semantic segmentation generated by SegDiscover with different initializations on images from COCO-Stuff dataset  (part 2). The column heading shows the initialization being used by SegDiscover.}
   \label{fig:app_segcom2}
\end{figure*}

Figure~\ref{fig:app_segcom1} and~\ref{fig:app_segcom2} show more results on the COCO-Stuff \cite{caesar2018coco} dataset. Here the class label for each of the 200 clusters was assigned to be the maximum class label within the cluster. 

\begin{table*}
    \centering
  \begin{tinyb}
  \begin{scriptsize}
  \caption{Classes in the COCO-Stuff dataset that are discovered by SegDiscover with the specified initialization.}
  \begin{tabular}{@{}l|ccccccccc}
    \hline
     Classes & electronic & appliance & food & furniture & indoor & kitchen & accessory & animal & outdoor \\
    \hline
Supervised&\checkmark  & & &\checkmark  &\checkmark  & & &\checkmark  &\checkmark  \\
MoCo & & & &\checkmark  &\checkmark  & & &\checkmark  &\checkmark \\ 

SwAV &\checkmark  & &\checkmark  &\checkmark  &\checkmark  & & &\checkmark  &\checkmark  \\
DeepClusterV2 & & & &\checkmark  &\checkmark  & & &\checkmark  &\checkmark \\
\hline
Classes & person & sports & vehicle & ceiling & floor & food-stuff & furniture & rawmaterial & textile \\
\hline

Supervised&\checkmark  &\checkmark  &\checkmark  & &\checkmark  &\checkmark  &\checkmark  &\checkmark  & \\
MoCo &\checkmark  &\checkmark  &\checkmark  & & &\checkmark  &\checkmark  &\checkmark  & \\
SwAV&\checkmark  &\checkmark  &\checkmark  & & &\checkmark  &\checkmark  &\checkmark  &\checkmark  \\
DeepClusterV2&\checkmark  &\checkmark  &\checkmark  & & &\checkmark  &\checkmark  &\checkmark  & \\

\hline
Classes & wall & window & building & ground & plant & sky & solid & structural & water \\
\hline

Supervised&\checkmark  &\checkmark  &\checkmark  &\checkmark  &\checkmark  &\checkmark  &\checkmark  &\checkmark  &\checkmark  \\
MoCo &\checkmark  &\checkmark  &\checkmark  &\checkmark  &\checkmark  &\checkmark  &\checkmark  &\checkmark  &\checkmark  \\
SwAV&\checkmark  &\checkmark  &\checkmark  &\checkmark  &\checkmark  &\checkmark  & &\checkmark  &\checkmark  \\
DeepClusterV2&\checkmark  &\checkmark  &\checkmark  &\checkmark  &\checkmark  &\checkmark  & &\checkmark  &\checkmark \\ 

\hline
    
  \end{tabular}
\end{scriptsize}
\end{tinyb}
  \label{tab:discoveredclasses}
\end{table*}

\clearpage
%
%
\bibliographystyle{splncs04}
\bibliography{segdiscover}